\documentclass[sigconf]{aamas} 

\usepackage{algorithm}
\usepackage[noend]{algorithmic}
\usepackage{color,soul}

\usepackage{amsfonts}
\usepackage{amsmath}
\usepackage{amsthm}
\usepackage{listings, array, blkarray}
\usepackage{graphicx}
\usepackage{bbm, dsfont}
\usepackage{multicol}
\usepackage{wrapfig,multirow}
\usepackage[caption=false]{subfig}
\usepackage{balance}

\newcommand{\matindex}[1]{\mbox{\scriptsize#1}}
\usepackage{comment}
\newtheorem{theorem}{Theorem}

\newtheorem{exmp}{Example}
\usepackage{comment}
\DeclareMathOperator*{\argmin}{argmin}
\usepackage{multirow}
\usepackage{wrapfig,lipsum,booktabs}

\usepackage{xcolor}

\newcommand{\LPlabel}{}

\newenvironment{LP}[3][]{%
\renewcommand{\LPlabel}{#1}
\ifthenelse{\equal{\LPlabel}{}}{%
\[ \begin{array}{ll}
\mbox{#2} & \;\;#3 \\
\mbox{subject to} & \begin{array}[t]{ll}%
}{%
\begin{equation} \begin{array}{ll}
\mbox{#2} & \;\;#3 \\
\mbox{subject to} & \begin{array}[t]{ll}%
}}{%
\ifthenelse{\equal{\LPlabel}{}}{%
\end{array} \end{array} \]}{%
\end{array} \end{array} \label{\LPlabel} \end{equation}}%
}

\usepackage{balance} 

\setcopyright{ifaamas}
\acmConference[AAMAS '22]{Proc.\@ of the 21st International Conference
on Autonomous Agents and Multiagent Systems (AAMAS 2022)}{May 9--13, 2022}
{Online}{P.~Faliszewski, V.~Mascardi, C.~Pelachaud,
M.E.~Taylor (eds.)}
\copyrightyear{2022}
\acmYear{2022}
\acmDOI{}
\acmPrice{}
\acmISBN{}

\acmSubmissionID{???}

\title[Networked Restless Multi-Armed Bandits for Mobile Interventions]{Networked Restless Multi-Armed Bandits for Mobile Interventions}

\author{Han-Ching Ou$^1$*, Christoph Siebenbrunner$^1$*, Jackson Killian$^1$, Meredith B Brooks$^1$, David Kempe$^2$, Yevgeniy Vorobeychik$^3$, Milind Tambe$^1$\\
$^1$Harvard University, $^2$University of Southern California, $^3$Washington University in St.~Louis
}
\email{{hou@g., csiebenbrunner@seas., jkillian@g., Meredith_Brooks@hms.}harvard.edu,}
\email{dkempe@usc.edu, yvorobeychik@wustl.edu, milind_tambe@harvard.edu}

\begin{abstract}
Motivated by a broad class of mobile intervention problems, we propose and study restless multi-armed bandits (RMABs) with network effects. In our model, arms are partially recharging and connected through a graph, so that pulling one arm also improves the state of neighboring arms, significantly extending the previously studied setting of fully recharging bandits with no network effects. 
In mobile interventions, network effects may arise due to regular population movements (such as commuting between home and work). We show that network effects in RMABs induce strong reward coupling that is not accounted for by existing solution methods. We propose a new solution approach for networked RMABs, exploiting concavity properties which arise under natural assumptions on the structure of intervention effects. We provide sufficient conditions for optimality of our approach in idealized settings and demonstrate that it empirically outperforms state-of-the art baselines in three mobile intervention domains using real-world graphs.
\end{abstract}


\newcommand{\BibTeX}{\rm B\kern-.05em{\sc i\kern-.025em b}\kern-.08em\TeX}

\begin{document}


\pagestyle{fancy}
\fancyhead{}


\maketitle 
\def\thefootnote{*}\footnotetext{The first two authors have equal contributions.}\def\thefootnote{\arabic{footnote}}


\section{Introduction}\label{sec:Introduction}
Mobile interventions are a model for providing services in which agents are sent to different locations where they provide various forms of interventions locally. Of particular importance are mobile health clinics (MHCs), a model of healthcare delivery in which mobile units deliver health services directly to target communities. MHCs are successful in reaching vulnerable populations; they overcome typical barriers to  health services access, such as limited transportation, finances, insurance, or legal status~\cite{stephanie2017scope}. A wide variety of MHC services---such as primary care, prevention screenings, disease management, and treatment support---have been very successful. Their success is based on their flexibility in meeting the changing needs of target communities, and providing these services at discounted rates or free of charge. Compared to other healthcare service models, MHCs have been observed to provide cost savings and cost-effectiveness~\cite{stephanie2017scope}. 
Another important application of mobile interventions is in food pantry services, which cater to communities experiencing food insecurity by dispatching food trucks.


Restless multi-armed bandits (RMABs) have become a widely adopted mathematical model for studying various types of intervention services \cite{kumar2010optimal,deo2013improving,mansour2015bayesian,lee2019optimal,mate2020collapsing,biswas2021learn,xu2021dual}. RMABs are a model for sequential planning problems: in each round, a planner has to select $k$ out of $m$ arms to pull. Arms transition randomly between states, but the transition probabilities differ based on whether an arm was pulled or not. The arms dispense rewards depending on their state. In our motivating applications, arms represent locations, $k$ may represent the budget (e.g., number of available MHC units), and rewards are the number of people positively affected by an intervention. In this paper, we extend existing RMAB models for interventions by considering network effects. 
Such network effects often arise due to individual commuting behavior: when an MHC visits one location, it provides interventions not only to people who reside there, but also to others who have traveled to this location (e.g., as a part of their routine work-related commuting). On the flip side, the same MHC may \emph{miss} people who have traveled to a different location.
Visiting one location may thus deliver an intervention to residents of multiple locations, giving rise to network effects. To the best of our knowledge,
we are the first to consider RMAB models with network effects.

Network effects lead to significant new challenges in the formal model. Common solution approaches for RMABs treat each arm as a Markov Decision Process (MDP) and exploit the fact that these MDPs are coupled only through the joint budget constraint. This weak coupling forms the basis for solutions based on index values, which are computed separately for each of the $m$ arms. Policies that select the $k$ arms with the highest indices can be shown to be asymptotically optimal for several domains~\cite{honda2010asymptotically,maillard2011finite,kaufmann2012thompson}. We show that the aforementioned network effects induce a stronger coupling between arms, making these solution approaches significantly less effective. The main contributions of our work are (1) we present a class of RMAB models with network effects suitable for modeling mobile intervention domains, (2) we present a solution approach for this class of problems and provide sufficient conditions for the optimality of our approach, and (3) we show empirically that our solution delivers superior performance compared to existing approaches across multiple domains. 


\section{Related Work}\label{sec:Literature}

In the most general setting, the RMAB problem is known to be PSPACE-hard to solve optimally \cite{papadimitriou1994complexity}. However, by exploiting the problem structure of certain restricted classes of RMABs, efficient algorithms have been derived, sometimes with performance guarantees. The most popular of these is the Whittle index policy \cite{whittle1988restless} which is asymptotically optimal for \emph{indexable} bandits \cite{weber1990index} and fast to compute if a closed form can be derived for the index. Many works are dedicated to proving the indexability of different RMAB subclasses and deriving closed-form or efficient approximations of the Whittle index \cite{glazebrook2006some,mate2020collapsing,hsu2018age,akbarzadeh2019restless}. Others have provided sufficient conditions for indexability \cite{nino2001restless} or developed expensive methods for computing policies with tighter reward bounds \cite{bertsimas2000restless,adelman2008relaxations}. However, all of these methods rely on the idea that the only factor coupling the arms are one or more budget constraints
which we refer to as the \emph{weakly coupled property}. 
Thus, previous RMAB methods will not be applicable for our work as the network effect strongly couples the states, actions, transitions, and rewards of neighboring arms.

In terms of applications, RMAB models have been widely used for scheduling problems, such as machine maintenance and repair~\cite{wang2002survey,abbou2019group,glazebrook2006some}. 
In these works, machines in factories are modeled as arms, and the goal is to find the optimal schedule to visit factories to maintain the machines. Other examples include anti-poaching patrol planning (\cite{qian2016restless} propose a RMAB framework in which arms are poaching targets, and playing an arm corresponds to a patrol) or recommendation systems (e.g., for music streaming \cite{zeng2016online,yi2017scalable}). Such problems also motivated the recharging bandit model~\cite{kleinberg2018recharging}.
In this model, each arm's reward is determined by a function of the time elapsed since the arm was last pulled. Implicitly, this resets the arm's reward to time 0 whenever the arm is pulled. When these functions are increasing and concave for each arm, \cite{kleinberg2018recharging} develop a concave program to solve the optimal frequency of pulling each arm; the program's value upper-bounds the value of an optimal schedule. Scheduling the arm then becomes a pinwheel scheduling problem~\cite{holte1989pinwheel}, and \cite{kleinberg2018recharging} use a rounding scheme to approximate the scheduling of arm pulls, while obeying the frequency restriction. 
We extend this setting by allowing the arms' rewards to be only \emph{partially} reset when the arm is selected, as well as by considering network effects.

In the public health domain, this paper's focus, \cite{mate2020collapsing} proposed collapsing bandits to improve medication adherence through interventions on patients. \cite{lee2019optimal} and \cite{ayer2019prioritizing} proposed RMABs for scheduling cancer screenings and hepatitis treatments, respectively. In~\cite{deo2013improving}, the closest RMAB application to ours, the authors model the resource allocation problem of delivering school-based asthma care for children. The most important difference between our work and theirs is that we consider network effects in the RMAB model.

\textbf{Related Work in Network Planning}
Sequential resource allocation problems on networks constitute another active area of research. Previous works have considered the non-restless setting, in which arms remain static when they are not pulled, such as influence maximization~\cite{vaswani2015influence,chen2016combinatorial}, or have studied the network effect on state transitions~\cite{fekom2019sequential, ou2021active} instead of on actions. To the best of our knowledge, ours is the first work to study RMABs with interventions that have network effects.

\section{Problem Formulation}\label{sec:RMAB}
\textbf{General RMABs}. RMABs are a generalization of the well-studied multi-armed bandit model with many real-world applications. There are $m$ arms $V=\{1, 2, \ldots, m\}$; each arm $v \in V$ can be in one of several states $s_{v,t} \in \mathcal{S}$ at any time step $t\in\mathbb{N}$. At any time step, the decision maker can pull up to $k$ arms. Each chosen arm $v$ transitions in a Markovian fashion according to a transition matrix $\mathbf{P}^a$ and yields a reward $r_v(s_{v,t}) \ge 0$ that depends only on the state of the arm $v$ at time $t$. In the restless setting, arms that are not chosen also transition, according to a different matrix $\mathbf{P}^p$. 
The elements $p^a_{s,s'}$ ($p^p_{s,s'}$) 
of the transition matrix capture the probability of transitioning from state $s$ to $s'$ when the arm is played (not played). Let $V_{a,t}$ denote the set of arms being played at time step $t$. The total reward of time step $t$ can be expressed as $R_t=\sum_{v \in V_{a,t}} r_{v,t}(s_{v,t})$. 
Each arm can be described as a two-action Markov Decision Process (MDP) $(\mathcal{S},\{0,1\},\mathcal{R},\mathcal{P})$. An action of $1$ denotes that the arm is played and $0$ that the arm is not played. Given the $m$ MDPs and their initial states, the goal of this work is to find a policy for playing a sequence of $k$ arms per round to maximize the average reward $\bar{R}=\lim_{T \to \infty} \frac{1}{T} \sum_{t=0}^{T} R_t$.%
\footnote{Another frequently considered reward criterion is the discounted reward $\sum_{t=0}^{\infty} \beta^t R_t$ with $0\leq \beta < 1$.}

\textbf{Networked RMABs for mobile interventions}. We consider a setting where each arm $v$ corresponds to a location which has a population $n_v\in\mathbb{N}$. The state $s_v \in \mathcal{S} = \{0,\dots,n_v\}$ of a location is the number of healthy individuals. Individuals can either be in a healthy or, more generally, ``good'' state $G$ or in a ``bad'' state $B$. Pulling an arm means visiting a location with a mobile intervention service, thereby exposing individuals at the location to the intervention. We thus consider the transition matrices for individuals, depending on whether they receive an intervention ($\mathbf{P}^{a}_{v}$) or not ($\mathbf{P}^{p}_{v}$): 
{\small
\begin{gather}\label{pvd_transition_matrix}
\mathbf{P}^{a}_{v} = \begin{blockarray}{ccc}
  & \matindex{$G$} & \matindex{$B$}\\
    \begin{block}{c[cc]}
      \matindex{$G$} & 1-p^{a}_{v,GB} & p^{a}_{v,GB}  \\
      \matindex{$B$} & p^{a}_{v,BG} & 1-p^{a}_{v,BG} \\
    \end{block}
  \end{blockarray} \;,
  \mathbf{P}^{p}_{v} = \begin{blockarray}{ccc}
  & \matindex{$G$} & \matindex{$B$}\\
    \begin{block}{c[cc]}
      \matindex{$G$} & 1-p^{p}_{v,GB} & p^{p}_{v,GB} \\
      \matindex{$B$} & p^{p}_{v,BG} & 1-p^{p}_{v,BG} \\
    \end{block}
  \end{blockarray}.
\end{gather}
}

The transition probabilities are the same for all individuals with the same home location. Below, we will consider travel by individuals, which may result in them being exposed to the intervention at a different location. We stress that even in that case, an individual with home location $v$ will transition according to the matrix $P_v$. This is because the characteristics of one's neighborhood are an important factor for one's health \cite{ross2001neighborhood}, keeping in mind the intended application domains of the model. We assume that the transition probabilities and the initial states are known, but the transitions are not observed. This is because while population-level health data can be monitored, this rarely happens in real time. We omit subscripts when they are clear from the context.

In order to account for network effects from commuting (or more general travelling) behavior, we define a probability distribution for individuals over locations. Let $w_{u,v}\in[0,1]$ denote the probability that an individual with home location $v$ is actually present in location $u$ at any given moment (or that an individual from location $v$ receives the intervention if location $u$ is visited; we assume that individuals are sampled uniformly). Individuals can only be in one location at any given time, implying that $\sum_{u\in V} w_{u,v}=1$. The matrix $\mathbf{W}\in[0,1]^{m\times m}$ with elements $w_{u,v}$ is the weighted adjacency matrix of the travelling network. Introducing the travelling network has two effects: 

\begin{enumerate}
    \item Not all individuals from location $v$ are exposed to an intervention that visits $v$. In expectation, only $n_v w_{v,v}$ individuals from location $v$ will receive the intervention (transition according to ${P}^{a}_{v}$) due to a visit at location $v$. This property is an important extension of the recharging bandits model \cite{kleinberg2018recharging}; in that model, it is assumed that each intervention fully ``resets'' the arm, i.e., puts all individuals into the good state. 
    \item Individuals from other locations receive the intervention when $v$ is visited. In expectation, $\sum_{u\in V\setminus \{v\}} n_u w_{v,u} $ individuals from other locations receive the intervention at $v$. 
\end{enumerate}

The total number of individuals reached in any location thus depends on whether other locations are visited, and we define the vector $\mathbf{a}_t\in\{0,1\}^m$, with at most $k$ elements equal to $1$, to represent all actions taken in round $t$. The vector of expected fractions of the populations at each location $v$ reached by an action vector $\mathbf{a}$ is given by $\hat{\mathbf{w}}(\mathbf{a})=\mathbf{W} \cdot \mathbf{a}$. Letting $\hat{w}_v$ denote the $v$-th entry of $\hat{\mathbf{w}}$, we also define the weighted average transition probabilities for a location $v$ as $\hat{\mathbf{P}}_v (\mathbf{a})=\hat{w}_v(\mathbf{a}_{t}) \cdot \mathbf{P}^a_v + (1-\hat{w}_v(\mathbf{a}_{t})) \cdot \mathbf{P}^p_v$. Further let $\mathbf{s}_{v,t} = [s_{v,t}, n_v-s_{v,t}]$ be the total number of individuals in the good and bad state in location $v$ at time $t$. By conditioning on the current state $\mathbf{s}_{v,t}$ and actions, we are able to obtain a closed form expression for the expected state in the next time step:
\begin{align*}
\mathbb{E}(\mathbf{s}_{v,t+1} \text{ } | \text{ } \mathbf{s}_{v,t}, \mathbf{a}_t,\dots,\mathbf{a}_0)
=&\mathbb{E}(\mathbf{s}_{v,t+1} \text{ } | \text{ } \mathbf{s}_{v,t}, \mathbf{a}_t)\\
=&\hat{w}_v \mathbf{s}_{v,t}  \mathbf{P}^a_v+(1-\hat{w}_v)\mathbf{s}_{v,t} \mathbf{P}^p_v  \\
=& \mathbf{s}_{v,t} \hat{\mathbf{P}}_v(\mathbf{a}_t).
\end{align*}

However, the current state is unknown according to our assumptions. Hence we seek an expression for the expected future state that does not require knowledge of the current state. Consider the expected state at time $t$ conditional only on the action history:
$\mathbb{E}_t(\mathbf{s}_{v,t}) := \mathbb{E}(\mathbf{s}_{v,t} \text{ } | \text{ } \mathbf{a}_{t-1},\dots,\mathbf{a}_0)$. Using the law of total expectation, we obtain
\begin{align*}
\mathbb{E}_{t+1}(\mathbf{s}_{v,t+1})=& \mathbb{E}(\mathbf{s}_{v,t+1} \text{ } | \text{ } \mathbf{a}_{t},\dots,\mathbf{a}_0)\\ 
=& \mathbb{E}(\mathbb{E}(\mathbf{s}_{v,t+1} \text{ } | \text{ } \mathbf{s}_{v,t}, \mathbf{a}_t) \text{ } | \text{ } \mathbf{a}_t,\dots,\mathbf{a}_0) \\
=& \mathbb{E}(\mathbf{s}_{v,t} \hat{\mathbf{P}}_v(\mathbf{a}_t) \text{ } | \text{ } \mathbf{a}_t,\dots,\mathbf{a}_0)\\ 
=& \mathbb{E}(\mathbf{s}_{v,t} \text{ } | \text{ } \mathbf{a}_{t-1},\dots,\mathbf{a}_0) \hat{\mathbf{P}}_v(\mathbf{a}_t),
\end{align*}
since $\mathbf{s}_{v,t}$ does not depend on $\mathbf{a}_t$ (only on previous actions). We thus obtain a recurrence relation for the expected state: 

\begin{equation}\label{eq:recurrence}
\mathbb{E}_{t+1}(\mathbf{s}_{v,t+1}) = \mathbb{E}_{t}(\mathbf{s}_{v,t}) \hat{\mathbf{P}}_v(\mathbf{a}_t).
\end{equation}

Eq.~\eqref{eq:recurrence} allows us to compute the future expected state using only the current expectation and action vector. In order to fully describe the probability distribution of a single district, one would need $\binom{m}{k}$ matrices of size $(n_v+1) \times (n_v+1)$. Eq.~\eqref{eq:recurrence} allows us to substantially reduce the complexity of the problem by focusing on the expected state. We write $\mathbb{E}_t(\mathbf{s}_{v,t})=\mathbf{b}_{v,t}$ and use the recursion $\mathbf{b}_{v,t+1}=\mathbf{b}_{v,t}\hat{\mathbf{P}}_v(\mathbf{a}_t)$, where the initial state $\mathbf{b}_{v,0} = \mathbf{s}_{v,0}$ is known according to our assumptions.

 The goal of the planner is to maximize the intervention benefit, taken as the sum of curing effects ($\text{cure}_v=p^{a}_{v,BG}-p^{p}_{v,BG}$) and prevention effects ($\text{prevention}_v=p^{p}_{v,GB}-p^{a}_{v,GB}$) for those individuals who received the intervention ($\text{cure}_v \hat{w}_v b_{v,t,2} + \text{prevention}_v \hat{w}_v b_{v,t,1}$, where $b_{v,t,1}$ and $b_{v,t,2}$ are the first and second element of $\mathbf{b}_{v,t}$, which are the expected total number of individuals in the good and bad state, respectively.), summed over locations and averaged over time steps. This criterion is chosen to align with the goals of applications such as MHCs which are to maximize the reach of a campaign \cite{auerbach20163}, and to avoid underserving communities with a high probability of returning to the bad state, as could happen if only the total number of people in the good state 
were considered. Combining the curing and prevention effects, the reward per time step is given by: $R_t(\mathbf{a}_t) = \sum_{v\in V}  \hat{w}_v(\mathbf{a}_{t}) \mathbf{s}_{v,t} (\mathbf{P}^a_v - \mathbf{P}^p_v) \cdot [1,0]^{\top}$. As discussed above, we focus on the expected reward and obtain:

\begin{equation}
\hat{R}_t := \mathbb{E}_t(R_t(\mathbf{a}_t) ) = \sum_{v\in V} \hat{w}_v(\mathbf{a}_{t}) \mathbf{b}_{v,t}  (\mathbf{P}^a_v - \mathbf{P}^p_v)\cdot [1,0]^{\top}.
\end{equation}

We further make three assumptions that are natural in many relevant application domains; we combine assumptions made in prior work~\cite{mate2020collapsing} (assumptions (1) and (2)) with input from health experts (assumption (3)). 

\begin{enumerate}
\item \textbf{The intervention is never bad for the individuals}: Health care interventions can help prevent disease or diagnose it early, reduce risk factors, and manage complications. Providing opportunities for increased access to quality services and interventions can reduce health disparities as well. Interventions provided via MHCs rarely result in negative impacts toward populations with little or no access to screening opportunities.  

\item \textbf{The individuals are more likely to stay in the good state than to change from the bad state to good}: In most applications, moving to the good state (curing of a disease or access to food) is unlikely to happen spontaneously. 

\item \textbf{The curing effect of the intervention is larger then the prevention effect}: MHCs mostly serve otherwise under-served communities. Those who attend MHCs are typically concerned about their health and may already be exhibiting symptoms of underlying disease. 
This makes curing interventions generally more useful/desired than preventive measures.
In food pantry applications, the prevention effect is typically small.
\end{enumerate}
These assumptions are formalized in Eq.~\eqref{eq:natural_assumptions}, for all $v\in V$:

\begin{subequations}\label{eq:natural_assumptions}
\begin{align}
p^{p}_{v,GB} &\geq p^{a}_{v,GB} \text{ and } p^{a}_{v,BG} \geq p^{p}_{v,BG} \\
1-p^{p}_{v,GB} &> p^{p}_{v,BG} \text{ and } 1-p^{a}_{v,GB} > p^{a}_{v,BG} \\
p^{a}_{v,BG}&-p^{p}_{v,BG} > p^{p}_{v,GB}-p^{a}_{v,GB}
\end{align}
\end{subequations}
Next, we show that these assumptions entail two properties that will prove useful later in constructing effective algorithms for the networked RMAB problem.
Specifically, consider a district $v$, and suppose that there are no interventions in adjacent districts.
We can then define the reward gain of visiting $v$ after $\tau_v$ time steps as $H^{\text{upper}}_v(\tau_v,\hat{w}_v)=(p^{p}_{v,GB}-p^{a}_{v,GB})\hat{w}_v \hat{s}_{v,\tau_v}+(p^{a}_{v,BG}-p^{p}_{v,BG})\hat{w}_v (n^{a}_{v,\tau_v}- \hat{s}_{v,\tau_v})$ where $\hat{s}_{v,\tau_v}$ is the number of individuals in the good state at the time when the arm pull happens. This function has the following properties:

\begin{theorem} \label{thm:concav}
Under the assumptions in Eq.~\eqref{eq:natural_assumptions}, and assuming no interventions in neighboring districts, $H^{u}_v$ is a monotone increasing concave function with respect to time $\tau_v$ elapsed since the last pull. 
\end{theorem}

\begin{theorem} \label{thm:portion}
Under the assumptions in Eq.~\eqref{eq:natural_assumptions}, and assuming no interventions in neighboring districts, $H^{u}_v$ is a monotone increasing concave function with respect to the expected population share $\hat{w}_v$ exposed to the intervention. 
\end{theorem}

The proofs are deferred to the supplementary material. Theorem~\ref{thm:concav} tells us that adding an extra pull to the intervention schedule of an arm will always improve the reward. From Theorem~\ref{thm:portion}, we know that it is always preferable to intervene on a larger proportion of the population of an arm.
These results suggest that the periodic policy is still a reasonable choice under the networked setting. The periodic policy in the non-networked setting is motivated by the following consideration: suppose that instead of pulling \emph{exactly} $k$ arms, we require only that \emph{on average}, $k$ arms are pulled in each round. In this relaxed problem, a periodic policy with suitable periods is optimal if the reward function is concave~\cite{kleinberg2018recharging}.\footnote{The constrained version is then a more difficult problem that involves solving a pinwheel problem, which is NP-hard.} Theorems~\ref{thm:concav} and \ref{thm:portion} tell us that the reward function for the networked problem is still concave. 

\section{Solution Approaches}\label{sec:algorithm}

As discussed previously, our problem shares significant similarities with the recharging bandits problem \cite{kleinberg2018recharging}. Both in the network-free and networked setting, a natural solution approach is to (1) determine the frequencies with which arms should be pulled, and then (2) sequence the pulls optimally. Importantly, the network effects affect both stages of the solution approach. As a result, simple optimal (or near-optimal) policies from the non-networked setting may be far from optimal when networks are considered.

The fact that network effects must be taken into account in determining arm pull frequencies is easy to see. Consider a star graph in which the central node has population 0, while the $m-1$ leaf nodes have population $n_v = n$, and --- importantly --- have probability 1 of commuting to the central node. Without considering the network/commuting effect, any policy would choose a non-central node in each round (because the central node has population 0), whereas picking the central node in each round is clearly optimal.

Perhaps more interestingly, network effects also impact which sets of arms should be pulled simultaneously, even keeping the arm pull frequencies constant (and having identical arms). This is illustrated in the following example.

\begin{figure}[ht]
\centering
\includegraphics[width=0.45\textwidth,keepaspectratio]{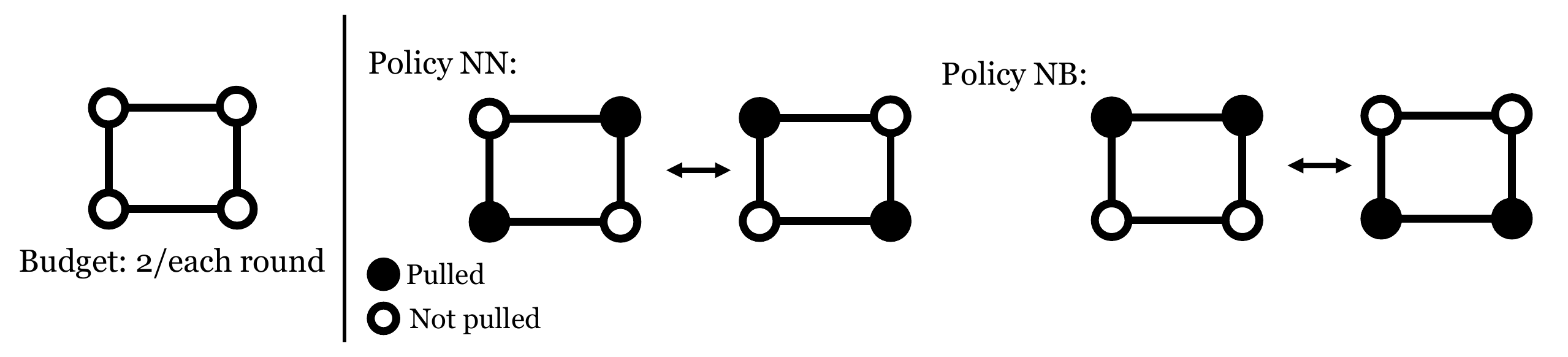}
\caption{Example for how network combinatorial effects affect the reward of periodic policies.}\label{fig:graph_example}
\end{figure}

\begin{exmp}\label{example:periodic}
Consider the example shown in Fig.~\ref{fig:graph_example}. We set $k=2$ and $(p^{p}_{GB},p^{p}_{BG},p^{a}_{GB},p^{a}_{BG})=(p_{GB},0,p_{GB},1)$.
All arms in Fig.~\ref{fig:graph_example} are identical. The optimal periodic policy is to select each arm every two rounds \cite{kleinberg2018recharging}. Such a policy can be achieved without any rounding by selecting exactly two arms in each round. However, different ways of choosing these two arms result in policies with different rewards.
Specifically, we consider the following two policies: Policy NN: Select two non-neighboring locations in each round. Policy NB: select two neighboring locations in each round. 
We also consider two different network scenarios with different commuting probabilities. In scenario 1, $w_{u,v}=\frac{1}{2}$ for all $(u,v) \in E$ and $w_{v,v}=0$ for all $v \in V$, i.e., all individuals commute to adjacent nodes. In scenario 2, $w_{u,v}=\frac{1}{4}$ for all $(u,v) \in E$ and $w_{v,v}=\frac{1}{2}$ for all $v \in V$, i.e., half of the individuals stay put.
Table~\ref{tab:square} summarizes the rewards of the two policies in the two scenarios: \begin{table}[htb]
    \centering
    \begin{tabular}{c|c|c}
    & Scenario 1 & Scenario 2 \\
    \hline
    Policy NN & $\frac{4p_{GB}-2p^2_{GB}}{1+p_{GB}-p^2_{GB}} \to 2$ & 
    $\frac{4 p_{GB}}{2p_{GB}+1} \to \frac{4}{3}$ \\
    Policy NB & $\frac{4 p_{GB}}{2p_{GB}+1} \to \frac{4}{3}$ &
    $\frac{52 p_{GB}-32p^{2}_{GB}}{13+16p_{GB}-16p^{2}_{GB}} \to \frac{20}{13}$
    \end{tabular}
    \caption{Rewards of the two policies, and limits as $p_{GB} \to 1$, in the two scenarios.}
    \label{tab:square}
\end{table}
In scenario 1, the policy NN is the better policy for any $p_{GB}$, and the relative reward difference can be as large as $\frac{2}{3}$.
In scenario 2, the policy NB becomes the better policy. For large $p_{GB}$, the relative reward difference approaches $\frac{13}{15}$. 
In particular, we see that the network effects must be taken into account in order to find the optimal way to coordinate the arm pulls of different arms.
\end{exmp}

Our proposed solution consists of two parts. In Section~\ref{sec:period}, we present an approach to obtain the optimal visiting period for each district. In Section~\ref{sec:sync}, we illustrate our approach for synchronizing the arm pulls to optimize reward coupling. 

\subsection{Proposed approach}\label{sec:proposed_solution}
Despite the added model complexities compared to the non-networked Recharging Bandits model, our problem preserves similar concavity properties. In a similar vein as \cite{kleinberg2018recharging}, we thus aim to provide periodic policies for the networked RMAB problem, i.e., policies that repeat after $T$ time steps. This not only facilitates scheduling, but can also reinforce intervention benefits in MHC domains \cite{willett2006prevention}. 
Exhaustively searching the action space of size $\binom{m}{k}^{T}$ is clearly impractical for reasonable problem sizes $m$. Fortunately, we can reduce the search space by exploiting the concavity we proved in Theorem~\ref{thm:concav}. 

\subsubsection{Obtaining Visiting Periods}\label{sec:period}
Let $x_v$ be the fraction of times that arm $v$ is chosen. When $1/x_v$ is integral, it can easily be shown that pulling the arm every $1/x_v$ rounds will maximize reward due to the concavity of the reward function~\cite{kleinberg2018recharging}. Define the period of pulling $\tau_v= 1/x_v\in\{1,2,3,\ldots,T\}$, meaning that $v$ is visited every $\tau_v$ time steps. Let $T$ be the maximum period considered, which could be a month, a season, or a year, depending on the application. Our goal is to find the optimal time period for each arm, subject to the sum of intervention frequencies being at most the budget $\sum_{v \in V} x_v \leq k$.

Suppose that a policy pulls arm $v$ every $\tau_v$ time steps and follows some schedule $\pi: t \rightarrow \mathbf{a}_t$. We define $\mathbf{P}^{*}_v(\tau_v,\pi)=\prod_{t=0}^{\tau_v}\mathbf{\hat{P}}_v(\pi(t))$ as the transition matrix of the expected state vector right before the next arm pull. Note that the reward gained from pulling an arm $v$ will depend on whether neighboring arms have recently been pulled, as this would imply that some share of $v$'s population has already been exposed to the intervention. For a given $\tau_v$, the reward gained from pulling $v$ is minimized when all neighboring arms are visited in every round and maximized when no locations other than $v$ are visited. We denote these two policies by $\pi^{\ell}$ and $\pi^{u}$, respectively. We can thus bound the average reward gained from pulling arm $v$ every $\tau_v$ rounds (defined as $H_v(\tau_v)$) as:
\[ \frac{1}{\tau_v}\mathbf{\bar{b}}^{\ell}_v\mathbf{P}^{*}_v(\tau_v,\pi^{\ell})\mathbf{n}_{v,G}\leq H_v(\tau_v) \leq \frac{1}{\tau_v}\mathbf{\bar{b}}^{u}_v\mathbf{P}^{*}_v(\tau_v,\pi^{u})\mathbf{n}_{v,G}, \]
where $\mathbf{\bar{b}}^{\ell}_v$ ($\mathbf{\bar{b}}^{u}_v$) is the steady state of $\mathbf{P}^{*}_v(\tau_v,\pi^{\ell})$ ($\mathbf{P}^{*}_v(\tau_v,\pi^{u})$), which is also its eigenvector corresponding to its smallest eigenvalue. $\mathbf{P}^{*}_v(\tau_v,\pi)$ is the $\tau_v$-step transition matrix of arm $v$ given the policy of other arms $\pi$.

Given the upper bound $H^{\text{upper}}_v(\tau_v)=\frac{1}{\tau_v}\mathbf{\bar{b}^{u}_v}\mathbf{P}^{*}_v(\tau_v,\pi^{u})\mathbf{n}_{v,G}$, we can construct the reward table for each arm $v$ by calculating the upper bound of each possible $\tau_v$. Finding the optimal period for each arm thus becomes an optimization problem
\[
\displaystyle \max  \sum_{v\in V} H^{\text{upper}}_v(\tau_v) \hspace{.5cm}
\textrm{s.t.} \sum_{v\in V} x_v \leq k.
\]



We explicitly write the optimization problem as a MILP with integer variables $x_{v,t} \in \{0,1\}$ for all $v \in V, t \in \{1, 2, \ldots, T\}$. $x_{v,t} = 1$ denotes that location $v$ has a period of $t$. In the MILP, we write $H^{u}_v(t) :=  \frac{1}{t}\mathbf{\bar{b}^{u}_v}\mathbf{P}^{*}_v(t,\pi^{u})\mathbf{n}_{v,G}$ for all $v$ and $t$.

\begin{LP}[LP:GetPeriodTable]{Maximize}{R}
\sum_{v} \sum_{t=1}^T \frac{x_{v,t}}{t} \leq k & \text{ (budget)}\\
\sum_{t=1}^T x_{v,t} \leq 1 \quad \text{ for all } v & \text{ (periods)}\\
R \leq \sum_{v\in V} \sum_{t=1}^{T} x_{v,t} H^{u}_v(t) & \text{ (reward)}\\
x_{v,t} \in \{0,1\} \quad \text{ for all } v, t.
\end{LP}

The MILP \eqref{LP:GetPeriodTable} has $O(|V|T)$ constraints. Its implementation can be found in the source code provided. The first constraint captures that the chosen periods/frequencies allow a fractional solution of at most $k$ visits per time step. The second set of constraints captures that each location has only one period. The third constraint bounds the reward.
From the MILP solution, for each $v$, the period $\tau_v$ can be obtained as the (at most one) $t$ such that $x_{v,t} = 1$.
If $x_{v,t} = 0$ for all $t$ for a particular $v$, then the arm is never worth pulling and can be discarded from the candidate pool.

The MILP can be adjusted to take fairness considerations into account as well. We list a few examples here; further details are discussed in the appendix:
\begin{itemize}
\item To achieve a minimum visiting frequency of $f_{\min}$, we can replace $T$ with $T_{\min} = 1/f_{\min}$.
\item To ensure that individuals from each node $v$ have sufficient access to the intervention (either at $v$ or a neighboring node), we can add the constraints $\sum_{u\in V} \sum_{t=0}^{T} \frac{w_{u,v} x_{u,t}}{t} \geq L$ for all $v$.
\item To encourage the algorithm to increase the smallest node rewards, we can replace the reward with the alternative welfare function $R \leq \sum_{v\in V} \sum_{t=1}^{T} x_{v,t} (\frac{H^{u}_v(t)}{n_v})^{\alpha}/\alpha$ for $\alpha\leq 1$.
\end{itemize}

\subsubsection{Finding optimal node sets to account for reward coupling}\label{sec:sync}
As illustrated in Example~\ref{example:periodic}, the combinatorial effects of pulling arms in the networked RMAB problem induce reward coupling between the MDPs of the arms. In contrast to non-networked recharging bandits, the choice of which set of arms with equal optimal periods to pull in the same rounds thus matters in networked bandits. The potential loss in reward here stems from the fact that when two arms that are both neighboring arms of a third arm are intervened on in different time steps, they will deliver the intervention in part to the same individuals in the third arm. 

In any time step $t$, for any pair of arms that is pulled simultaneously, we seek to maximize the overlap between the shares of populations in the set of arms that are neighbors of both arms. For a pair of arms $(v,v')$, this intervention overlap can be computed as $\sum_{u \in \delta(v) \cap \delta(v')} w_{v,u} w_{v',u}$. If (and only if) the optimal periods $\tau_v$ and $\tau_{v'}$ are coprime to each other, this intervention overlap is independent of when the arms are intervened on. (As an example, two arms with periods $2$ and $3$ will be pulled together every six rounds, regardless of when the policy starts pulling each arm.) If the periods $\tau_u$ and $\tau_v$ have a common factor, on the other hand, they can never be pulled together if they are out of sync. (Arms with periods $2$ and $4$ will never be pulled together if their sequences start one time step apart.) We would thus be losing out on the reward gains from pulling the arms together every $\text{lcm}(\tau_u,\tau_v)$ rounds. In order to minimize this loss, we construct an undirected graph $\bar{G}(V,\bar{E})$ with the following edge weights:




\begin{equation}\label{eq:G_bar}
\bar{w}_{v,v'}(\tau_v,\tau_{v'}) =
\begin{cases}
\sum_{u \in \delta(v) \cap \delta(v')} \frac{w_{v,u}w_{v',u}}{\text{lcm}(\tau_v,\tau_{v'})} 
     & \mbox{ if } \text{gcd}(\tau_v, \tau_{v'}) > 1 
\\ 0 & \mbox{ otherwise}
\end{cases}
\end{equation}

The weight of the cut between the selected and unselected arms on $\bar{G}$ equals the average reward loss due to the intervention overlap. We can thus select the arm set to pull by minimizing the cut between the selected node set (of size $k$) and the unselected node set. Graph partition problems with node cardinality constraints are generally NP-hard \cite{vazirani2013approximation}. We use a heuristic based on spectral graph partitioning, by considering the $k$ nodes with the largest or smallest value in the eigenvector corresponding to the second-smallest eigenvalue of $\bar{L}$ (also known as the Fiedler vector), where $\bar{L}$ denotes the Laplacian of the graph $\bar{G}$. The \textsc{ENGAge} (Efficient Network Geography Aware scheduling) Algorithm (Algorithm~\ref{alg:scheduling}) outputs an intervention policy based on this approach.


\begin{algorithm}[h]
\caption{\textsc{ENGAge} }\label{alg:scheduling}
\begin{flushleft}
\end{flushleft}
\begin{algorithmic}[1] 
\STATE $V_{\text{candidate}} \gets V$ and $V_{\text{wait}} \gets \emptyset$.
\STATE Compute periods $\tau_v$ using the MILP \eqref{LP:GetPeriodTable}.
\STATE Construct the new graph $\bar{G}(V,\bar{E})$ according to Eq.~\eqref{eq:G_bar} and compute its Laplacian $\bar{L}$. 
\STATE Find the set $\Lambda$ of Fiedler vectors of $\bar{L}$ (more than one in case of eigenvalue multiplicity).
\FOR{$t = 1,\dots,T$} \label{ln:for_start}
\FOR{$v \in V_{\text{wait}}$} \label{ln:for_start}
\STATE $\text{Timer}(v) \gets \text{Timer}(v)-1$.
\IF{$\text{Timer}(v)=0$}
    \STATE Move $v$ from $V_{\text{wait}}$ to $V_{\text{candidate}}$.
\ENDIF
\ENDFOR
\STATE $V_a(t) \gets \emptyset$.
\FORALL{$\eta \in \Lambda$}
\STATE Find the sets of nodes with $k$-th largest and smallest elements in $\eta$:
Specifically, let $\eta_{(k)}$ denote the $k$-th largest entry of $\eta$, set
$\bar{V} \gets \{v \in V_{\text{candidate}} \; | \; \eta_{v} \le \eta_{(k)} \}$ and $\underline{V} = \{v \in V_{\text{candidate}} \; | \; \eta_{v} \geq \eta_{(m-k+1)} \}$.
\STATE If $|\bar{V}|>k$ or $|\underline{V}|>k$, reduce the set size to $k$ by arbitrarily removing tied nodes at the cutoff threshold.
\STATE Update $V_a(t)$ to the set $S$ that minimizes the cut: 
$V_a(t) \gets \argmin_{S \in \{\bar{V}, \underline{V}, V_a(t)\}} c(S)$.
Here, $c(S)$ denotes the cut capacity of the node set $S$ in $\bar{G}$ (and is defined as $\infty$ for the empty set). Arbitrarily break ties. 
\ENDFOR
\STATE Move $V_a(t)$ from $V_{\text{candidate}}$ to $V_{\text{wait}}$, and set $\text{Timer}(v) \gets \tau_v$ for these arms.
\ENDFOR 
\RETURN $V_a(t)$ as arms to pull at time $t$ for all times $t = 1,\dots,T$.
\end{algorithmic}
\end{algorithm}

\begin{figure*}[ht]
\centering
\null\hfill
\subfloat[MHCs in urban area]{%
  \includegraphics[width=0.3\textwidth,keepaspectratio]{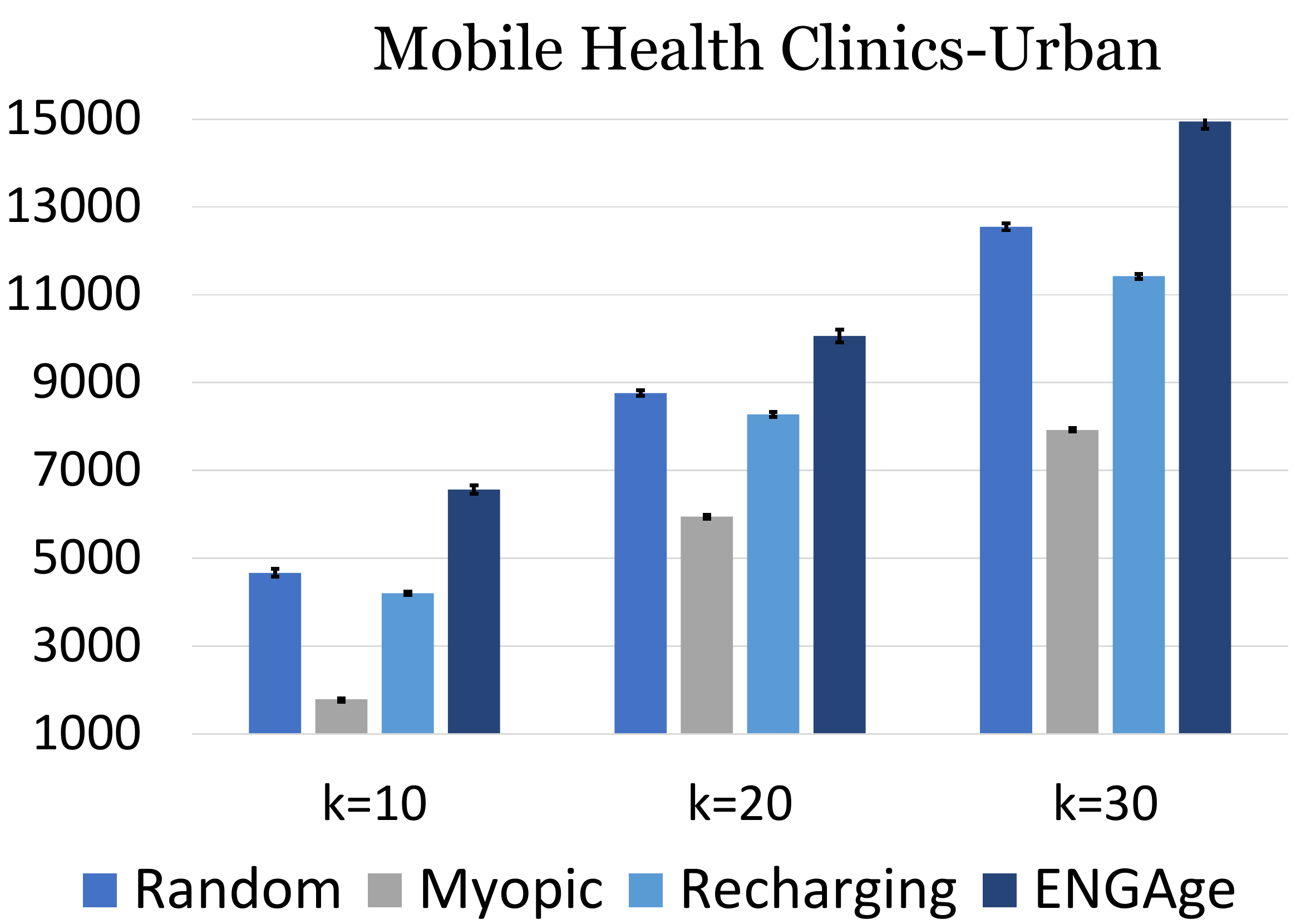}\label{fig:MHC}%
}\hfill
\subfloat[MHCs in rural area]{%
  \includegraphics[width=0.3\textwidth,keepaspectratio]{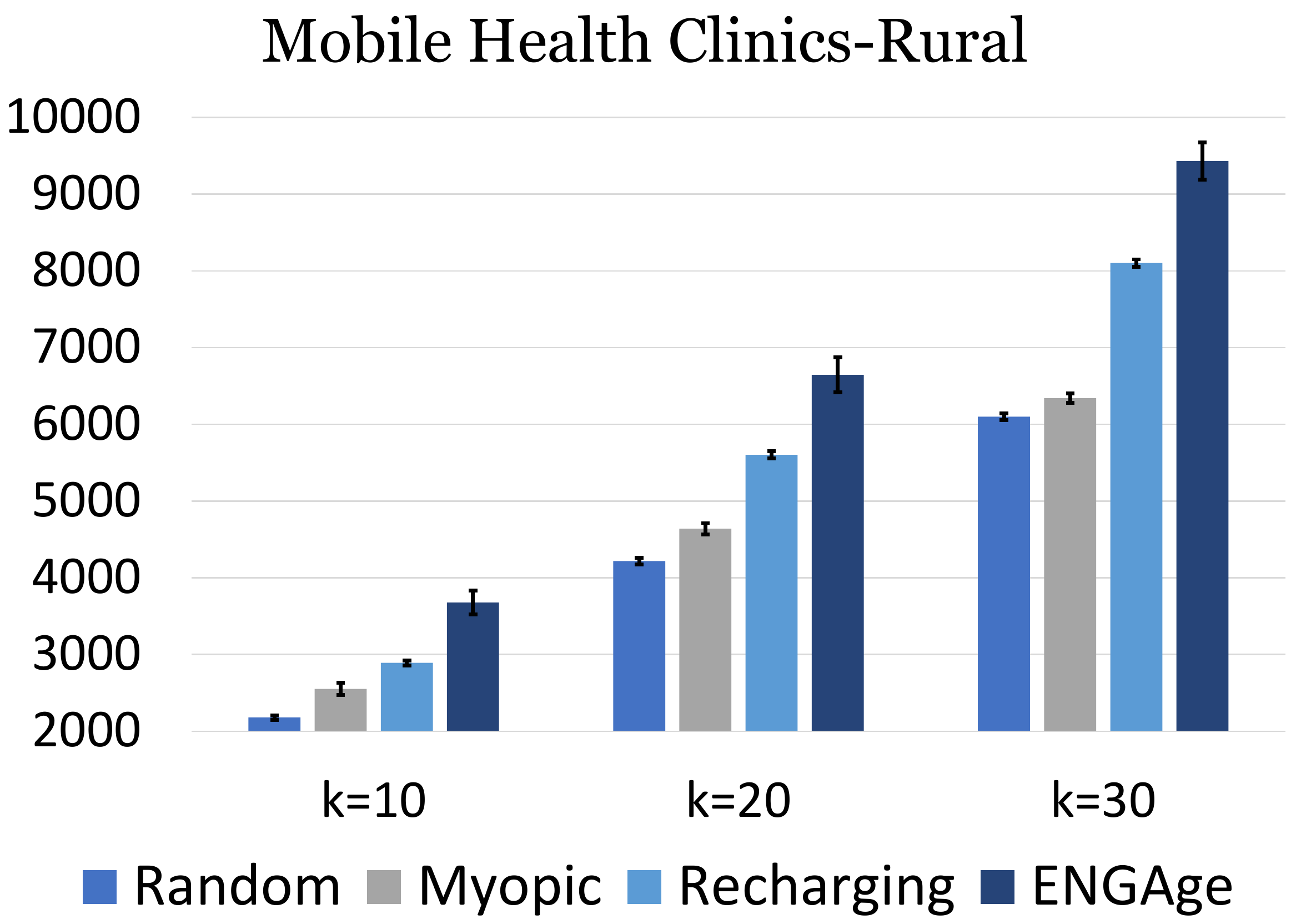}\label{fig:green_security}%
}\hfill
\subfloat[Mobile Food Pantry]{%
  \includegraphics[width=0.3\textwidth,keepaspectratio]{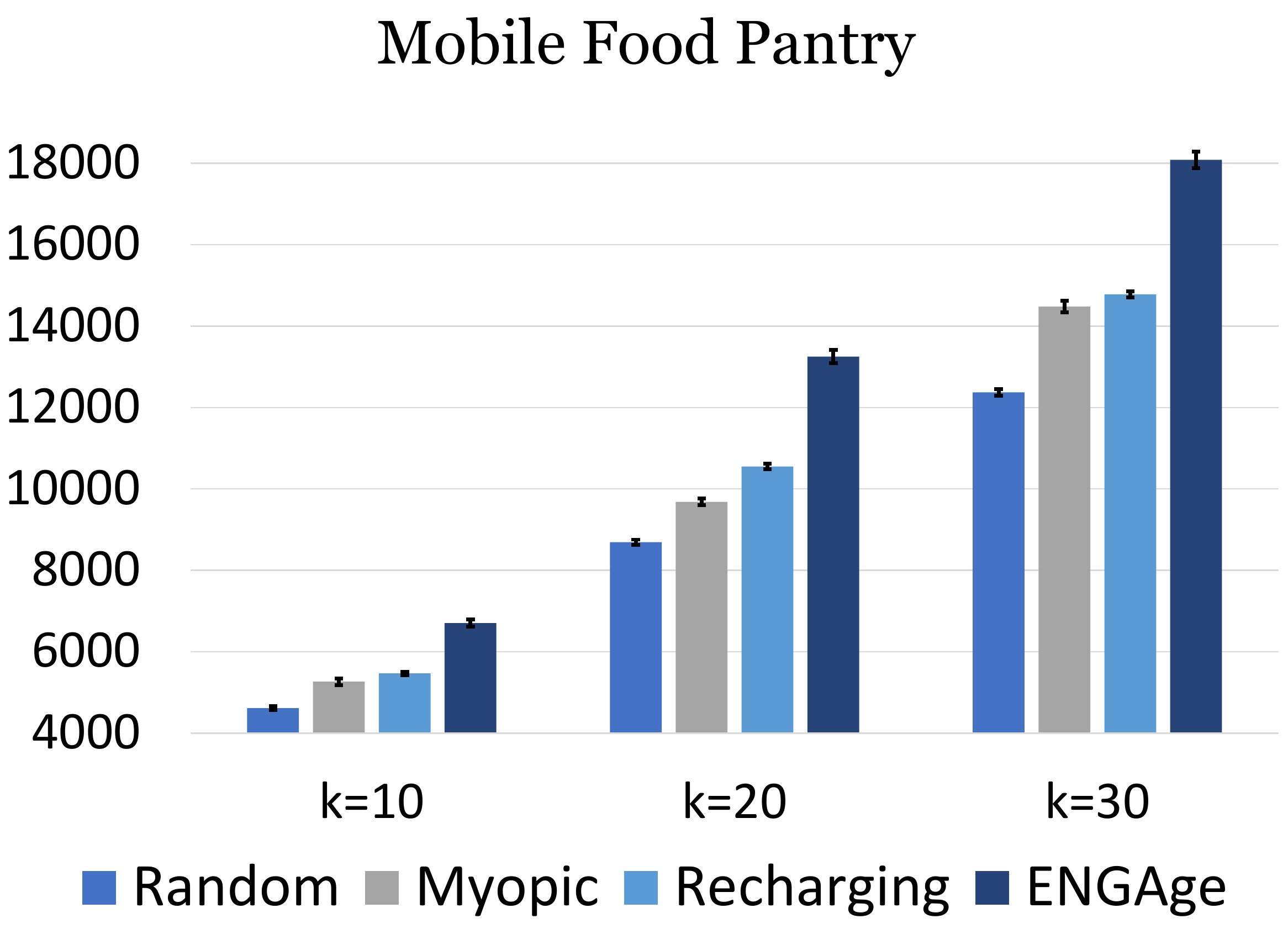}\label{fig:food}%
}\hfill\null
\centering
\caption{Average reward in three different domains under different budget constraints.}\label{fig:exp_result}
\end{figure*}

\begin{figure*}[ht]
\centering
\subfloat[Reward for different curing rates in urban MHC domain]{%
  \includegraphics[width=0.2\textwidth,keepaspectratio]{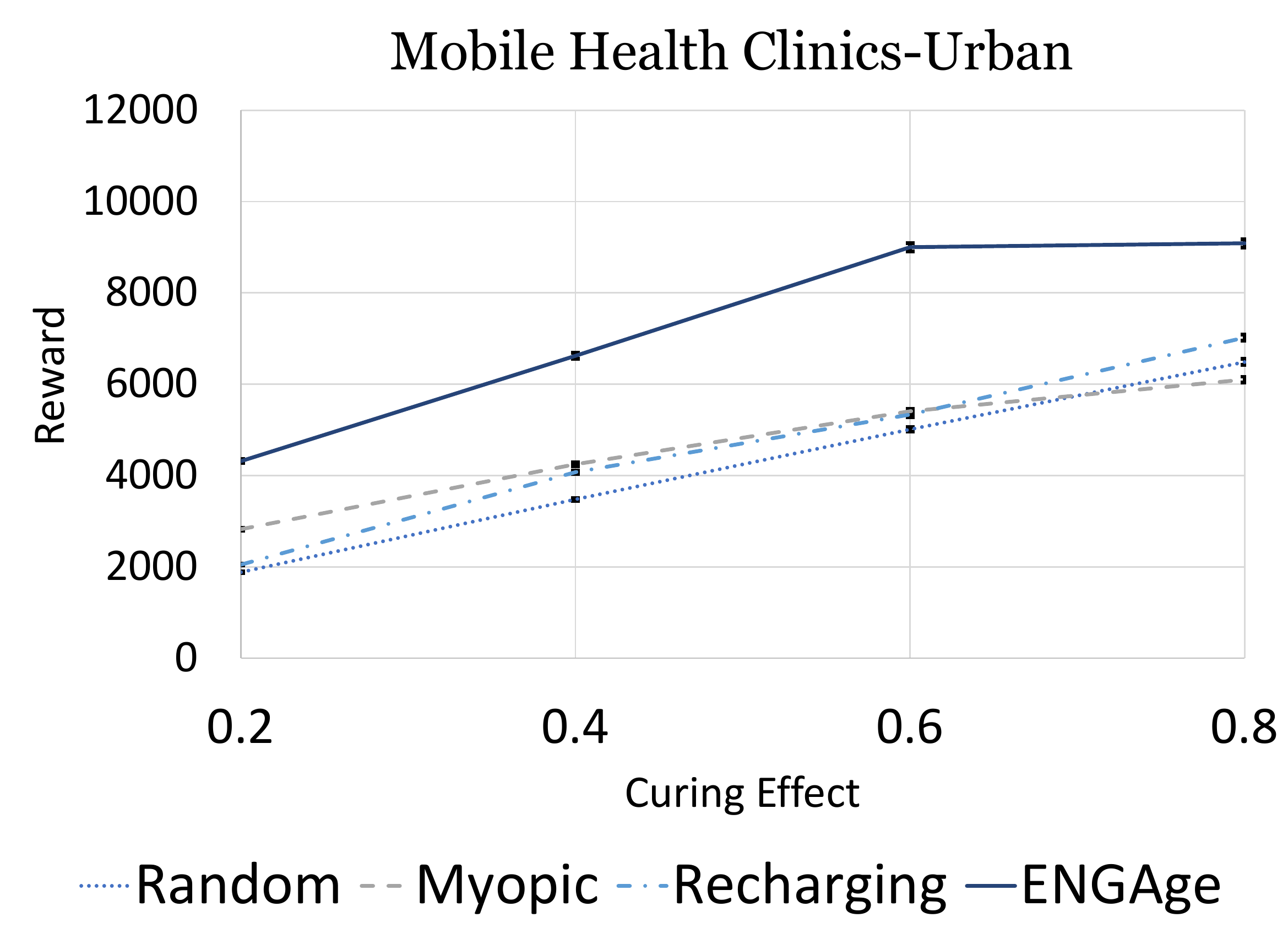}\label{fig:cure_MHC}%
}\hfill
\subfloat[Reward for different curing rates in rural MHC domain]{%
  \includegraphics[width=0.2\textwidth,keepaspectratio]{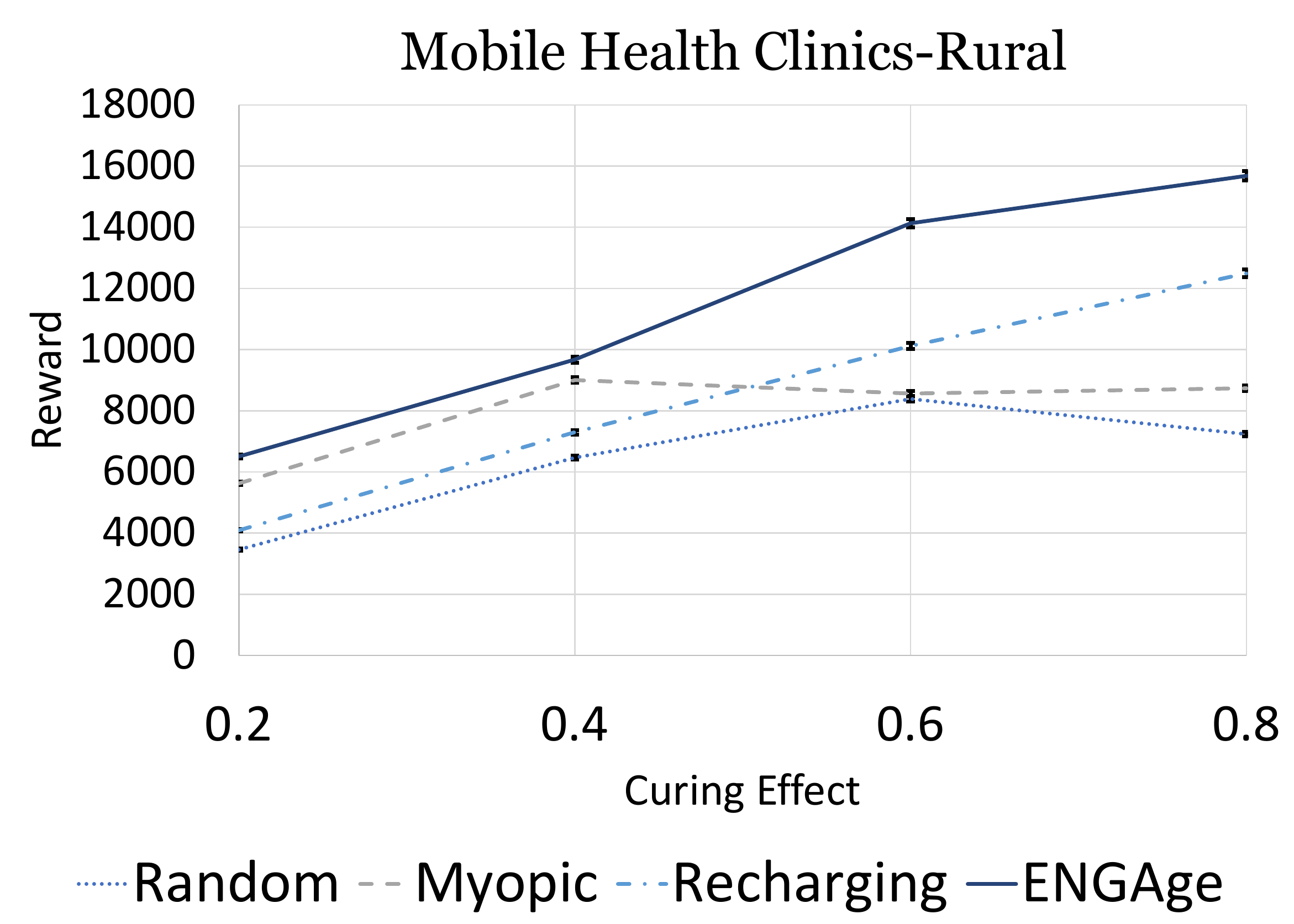}\label{fig:cure_rural}%
}\hfill
\subfloat[Reward for different curing rates in food pantry domain]{%
  \includegraphics[width=0.2\textwidth,keepaspectratio]{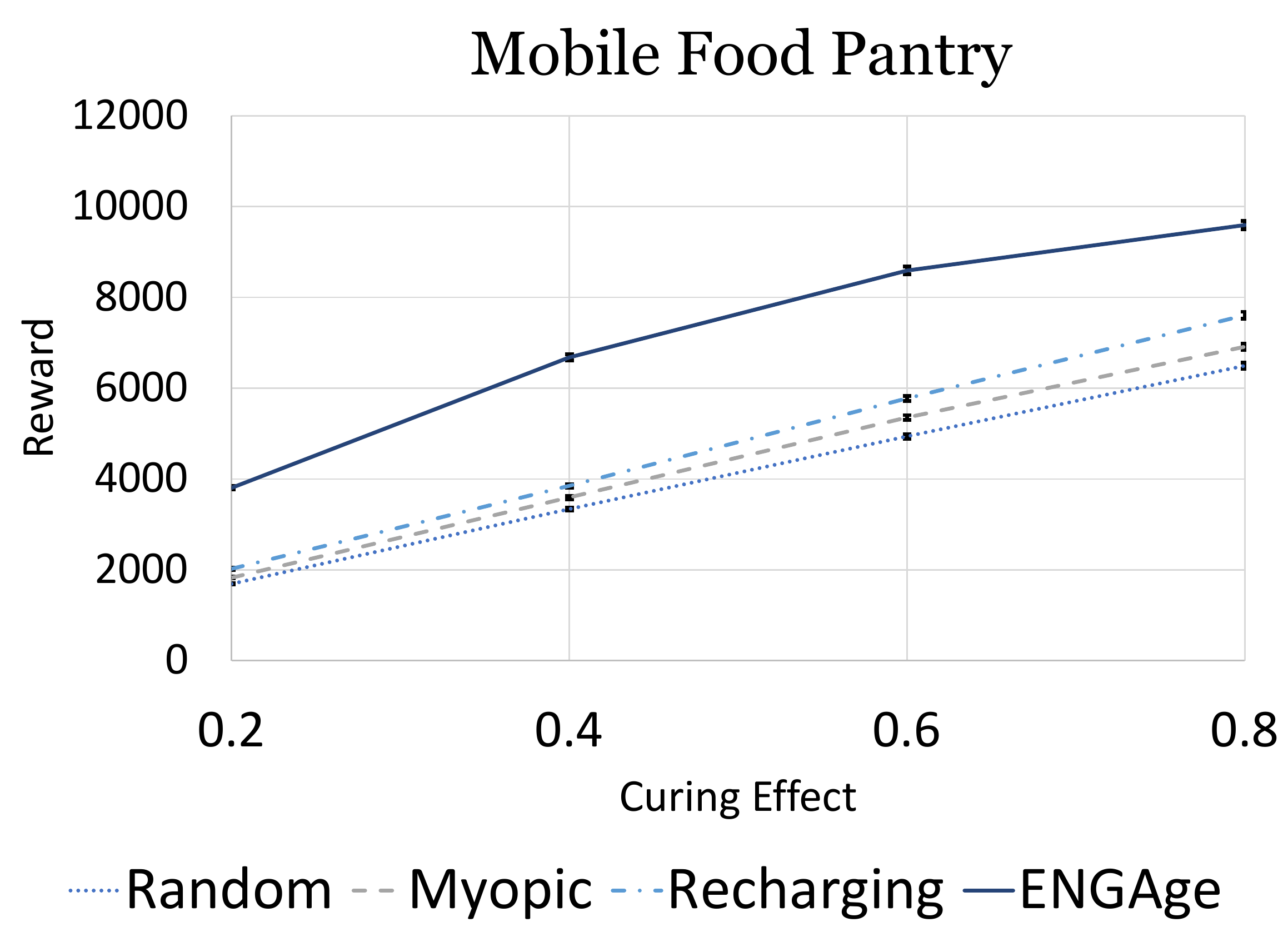}\label{fig:cure_food}%
}\hfill
\subfloat[Reward for different curing rates in synthetic domain]{%
  \includegraphics[width=0.2\textwidth,keepaspectratio]{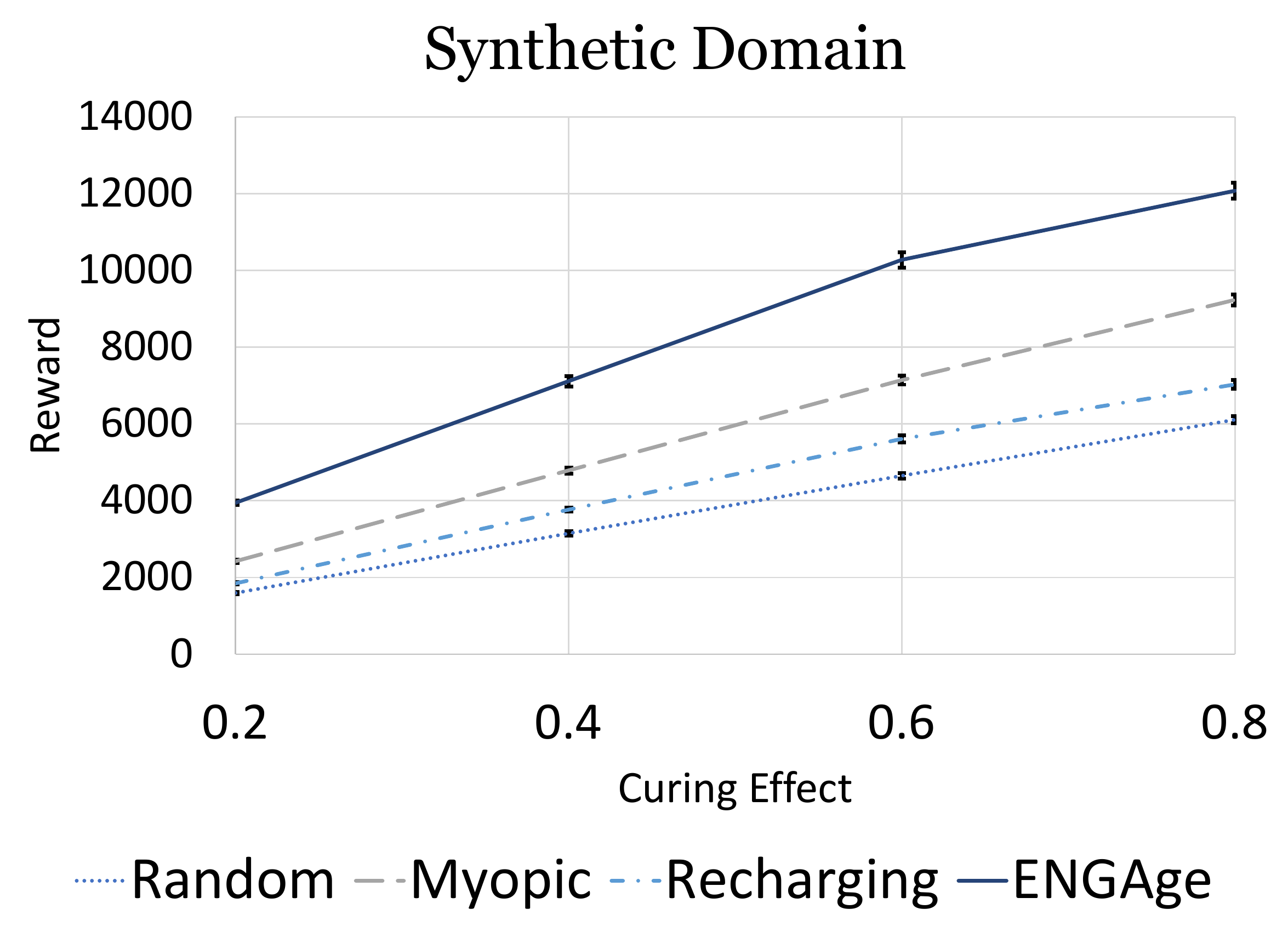}\label{fig:cure_syn}%
}
\centering
\caption{Average reward in three different domains under different curing effects ($P^{a}_{GB}-P^{p}_{GB}$) .}\label{fig:cure_all}
\end{figure*}

\begin{figure*}[ht]

\subfloat[Reward for prevention rates in urban MHC domain]{%
  \includegraphics[width=0.2\textwidth,keepaspectratio]{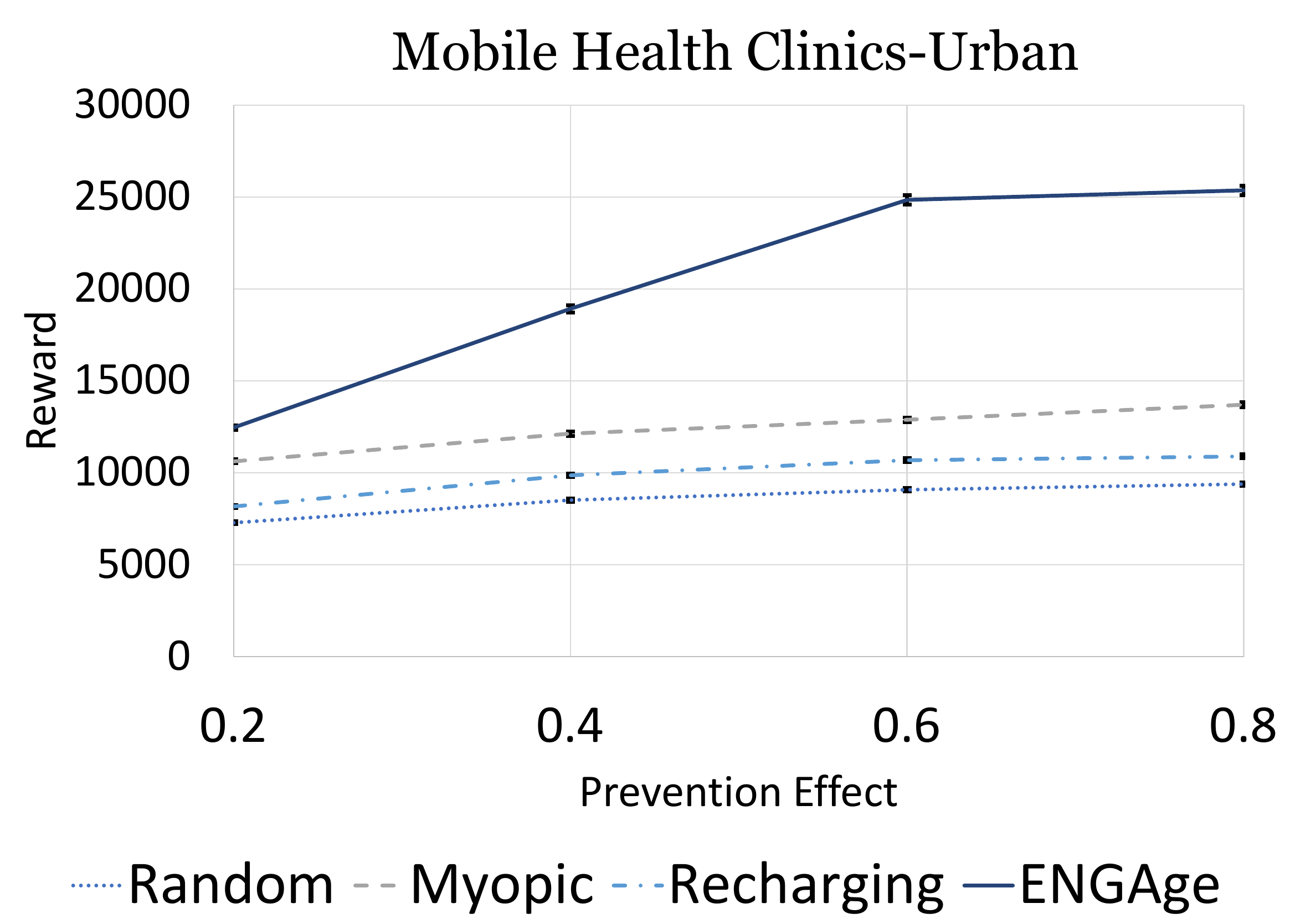}\label{fig:prevenrion_city}%
}\hfill
\subfloat[Reward for prevention rates in rural MHC domain]{%
  \includegraphics[width=0.2\textwidth,keepaspectratio]{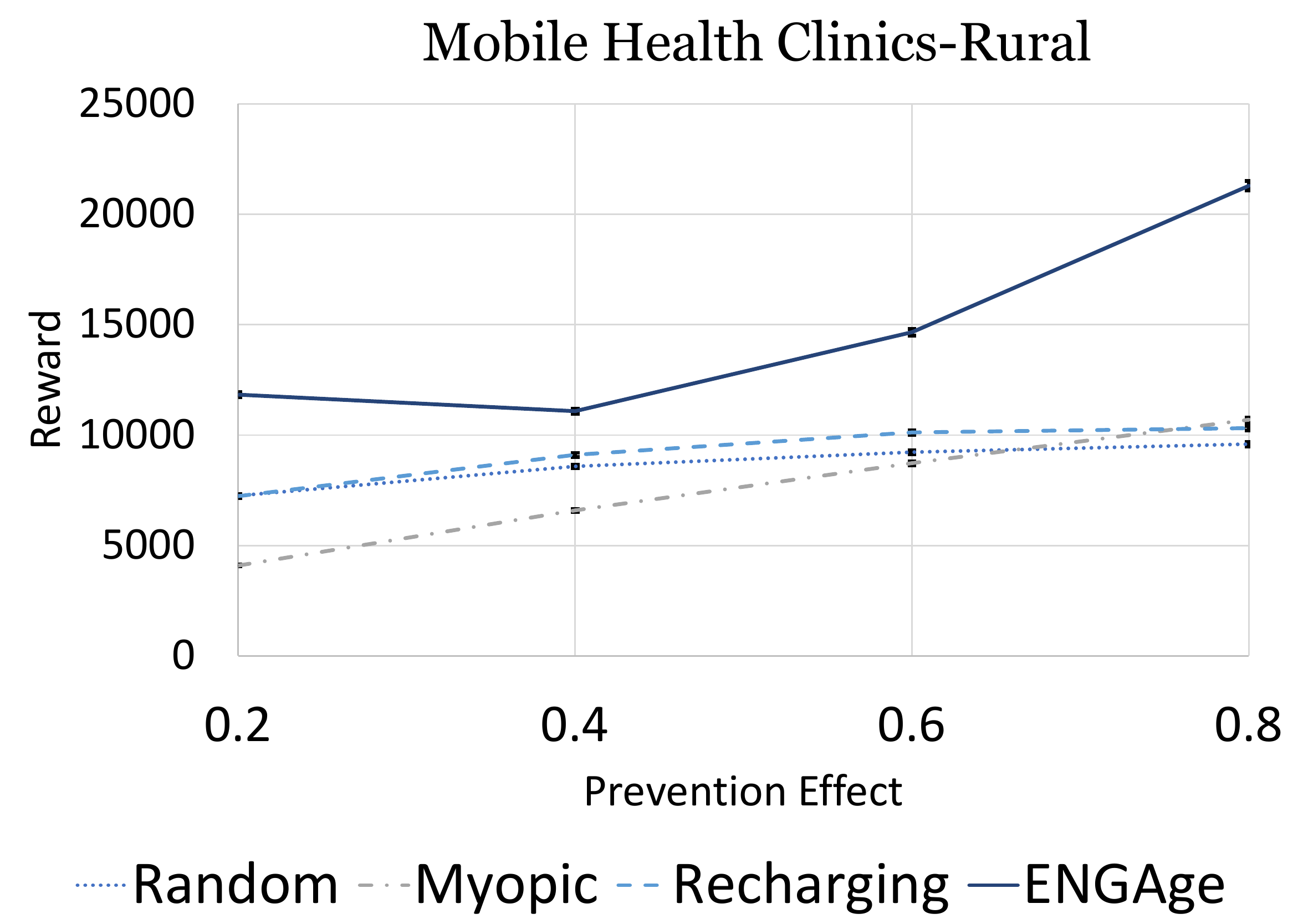}\label{fig:prevenrion_rural}%
}\hfill
\subfloat[Reward for prevention rates in food pantry domain]{%
  \includegraphics[width=0.2\textwidth,keepaspectratio]{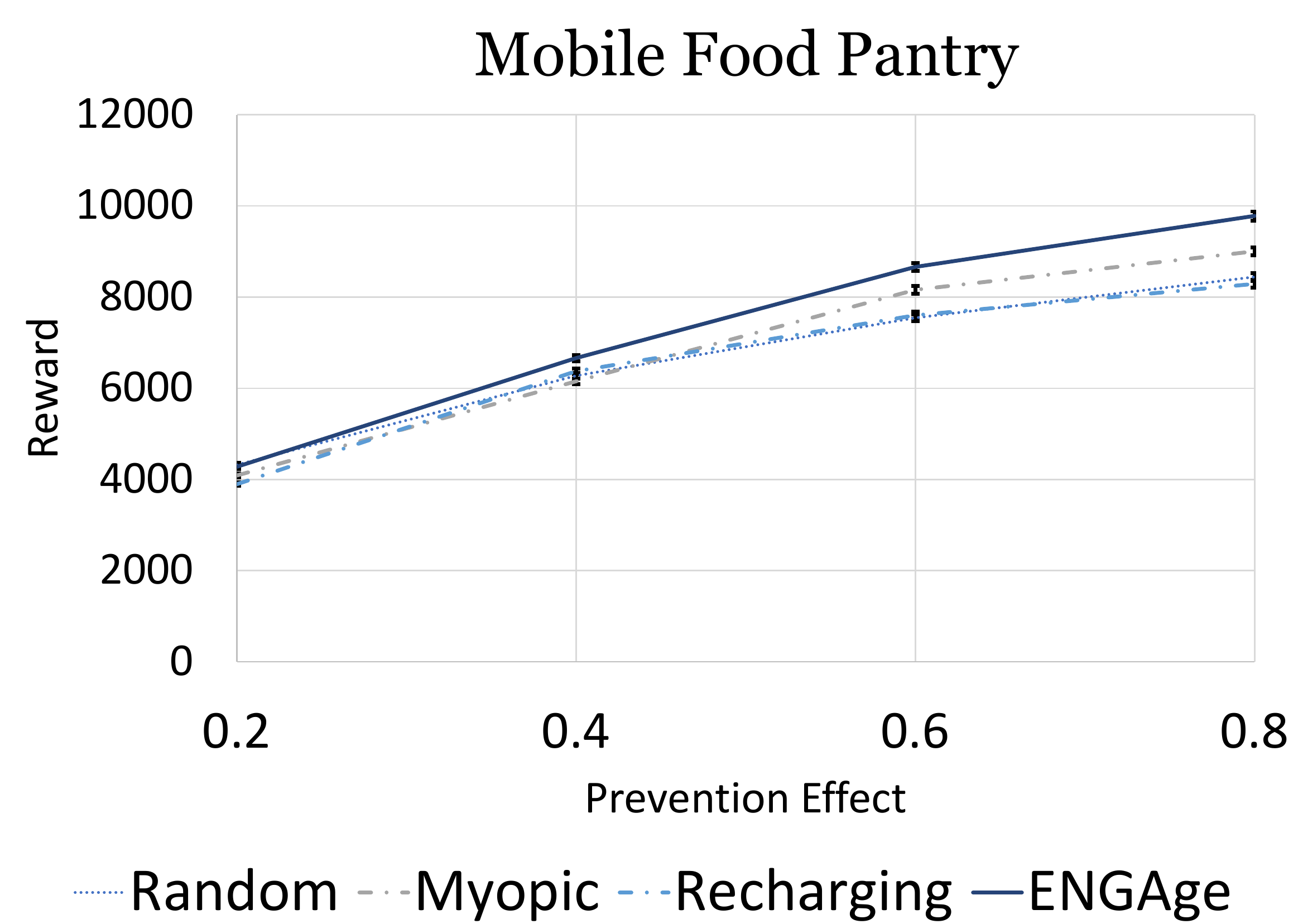}\label{fig:prevenrion_food}%
}\hfill
\subfloat[Reward for prevention rates in synthetic domain]{%
  \includegraphics[width=0.2\textwidth,keepaspectratio]{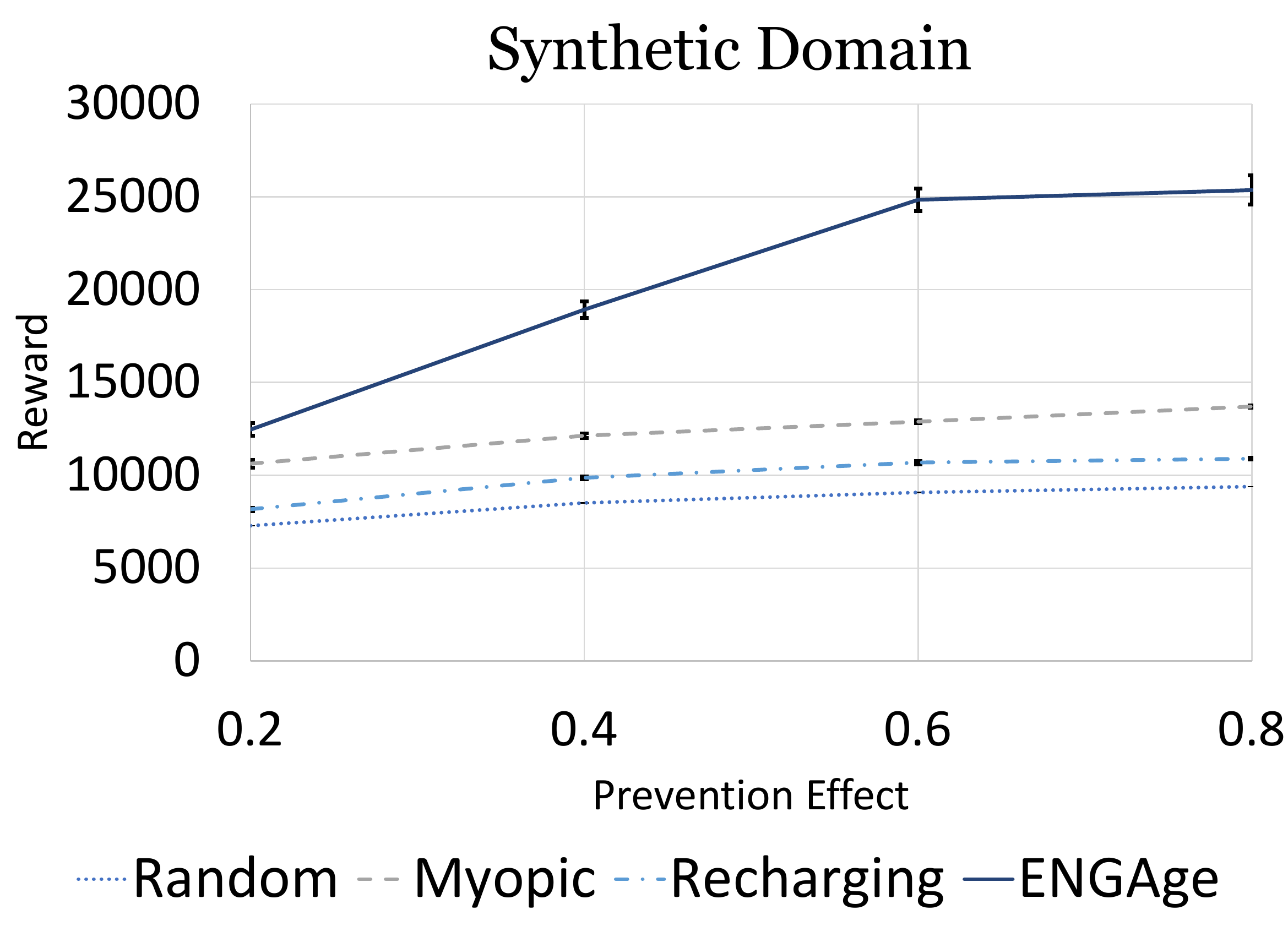}\label{fig:prevenrion_syn}%
}
\centering
\caption{Average reward in three different domains under different prevention effects ($P^{p}_{BG}-P^{a}_{BG}$) .}\label{fig:prevenrion_all}
\end{figure*}

\subsection{Analysis}\label{sec:analysis}

We start by analyzing the complexity of the solution approach described above. The concave MILP~\eqref{LP:GetPeriodTable} can be solved efficiently using time $O(|V|T \log (|V|T))$, by sorting the set of slopes of segments, corresponding to the different $H^{u}_{v}(t)$. Details are given in \cite{kleinberg2018recharging}. In our implementation, we instead use an off-the-shelf MILP solver. While its worst-case running time is larger, as our experiments show, it runs very efficiently in practice. Calculating the Laplacian $\bar{L}$ requires finding common neighbours ($O(\hat{d}|E|)$ by~\cite{an2019skip}, where $\hat{d}$ is the maximum degree in $\bar{G}$) and then computing their gcd ($O(\log T)$ using the Euclidean algorithm). The overall cost of computing the Laplacian is thus $O(\hat{d}|E|\log T)$. Again, our actual implementation is less efficient in terms of worst-case complexity, but runs fast in practice nonetheless. Finding a Fiedler vector takes time $O(\bar{d}|V|)$ using Lanczos' algorithm~\cite{lanczos1950iteration}, where $\bar{d}$ is the average degree of $\bar{G}$.
The rest of the planning takes time $O(|V|T)$. 
Thus, the total time complexity of our algorithm is $O(|V|T \log (|V|T) +\hat{d}|E|\log T)$.


In Section~\ref{sec:exp}, we experimentally evaluate the performance of our algorithm on various graphs from real-world domains. We now turn to analyzing sufficient conditions that guarantee optimality for various cases that we will discuss below. 

First consider the case of homogeneous nodes and edge weights, i.e., all nodes have the same populations and transition probabilities between states, and all edges have the same commute probabilities. If we replace the eigenvector-based heuristic in \textsc{ENGAge} with an oracle that optimally solves the min-cut problem with cardinality constraints, then \textsc{ENGAge} outputs the optimal policy for arbitrary graphs of $N$ nodes whenever $k | N$. This is because in this case, the cut on the constructed graph measures the exact reward loss of the schedule. Solving the min-cut problem optimally will then lead to the optimal scheduling.

Next, consider the special case in which the graph $\bar{G}$ has $\gamma$  connected components $C_1, \dots, C_\gamma$, each of size $|C_i|=k$. 
Furthermore, we assume that all elements of the same component have the same optimal period; that is, if $u, v \in C_i$, then $\tau_u = \tau_v$.
For $\gamma \ge 2$, note that $\bar{L}$ is positive semidefinite as $\bar{G}$ is undirected for arbitrary input graphs $G$ by construction.
The smallest eigenvalue $0$ will have multiplicity $\gamma$ in the Laplacian $\bar{L}$.
Thus, $|\Lambda|=\gamma$, and it is known that each component $C_i$ has a corresponding Fiedler vector supported entirely on $C_i$ \cite{marsden2013eigenvalues}. 
Hence, in each iteration, Algorithm~\ref{alg:scheduling} will select exactly all members of one component. As there are no links between nodes in different components by definition, all members of a component will be fully intervened on. Our problem thus reduces to a pinwheel problem with $\gamma$ arms and optimal periods $\tau_i$ for $i = 1,\dots,\gamma$. Pinwheel problems are known to be NP-hard in general~\cite{chan1993schedulers}, but optimal solutions are known to exist in special cases where all periods are multiples of one another and $\sum_i^\gamma \tau_i\le k$~\cite{holte1989pinwheel}. The optimal solution in these cases can be obtained by a simple greedy policy (see~\cite{chan1993schedulers}) which is realized by the sets $V_{\text{wait}}$ of our algorithm.
The latter condition is guaranteed by the setup of \textsc{ENGAge}; hence, our proposed approach will output an optimal schedule in those cases. For $\gamma=1$, the same conclusion follows trivially, because the algorithm can visit all locations in each time step.

Based on the above analysis, \textsc{ENGAge} will output the optimal policy in the following settings, among others: (1) Complete graphs with equal edge weights, identical nodes, and $k | N$. (2) Graphs with multiple connected components, each of size $k$, with equal edge weights and identical nodes. (3) Rings with edge weights  $1/2$, identical nodes, and $k=N/2$. (4) $d$-dimensional Hypercubes with edge weights $1/d$, identical nodes, and $k=N/d$. (5) Bipartite or multipartite graphs with partitions of size $k$, identical node degrees, and edge weights summing to $1$ for all nodes. (6) Strongly $d$-regular graphs with equal edge weights and identical nodes. These are illustrative examples of graphs where our algorithm is guaranteed to perform optimally. In the next section, we will empirically show that it outperforms existing methods in more general settings, including real-world graphs.

\begin{figure*}[ht]
\centering
\hfill
\subfloat[Average Degree]{%
  \includegraphics[width=0.3\textwidth,keepaspectratio]{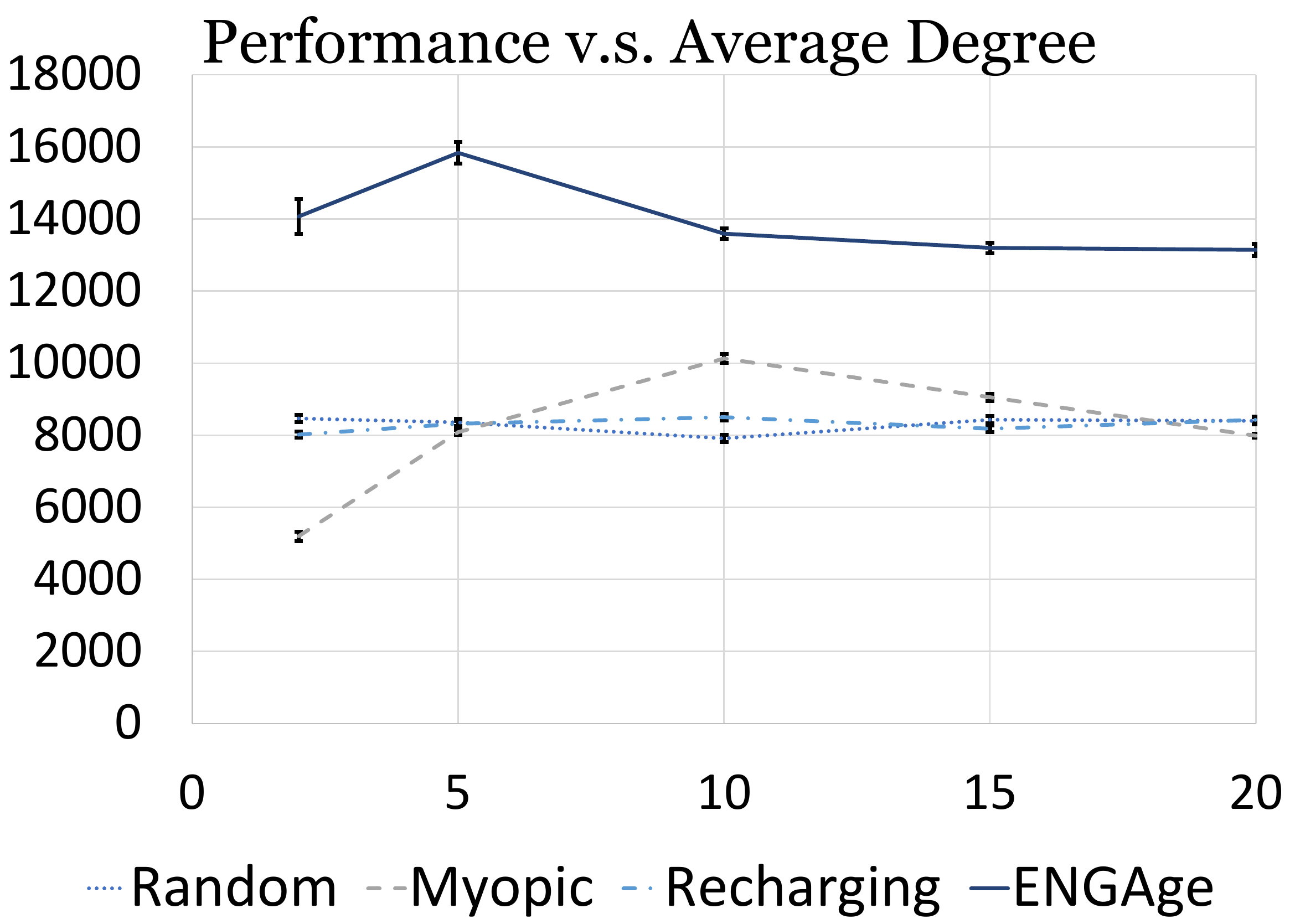}\label{fig:average_degree}%
}\hfill
\subfloat[Disadvantaged Communities]{%
  \includegraphics[width=0.3\textwidth,keepaspectratio]{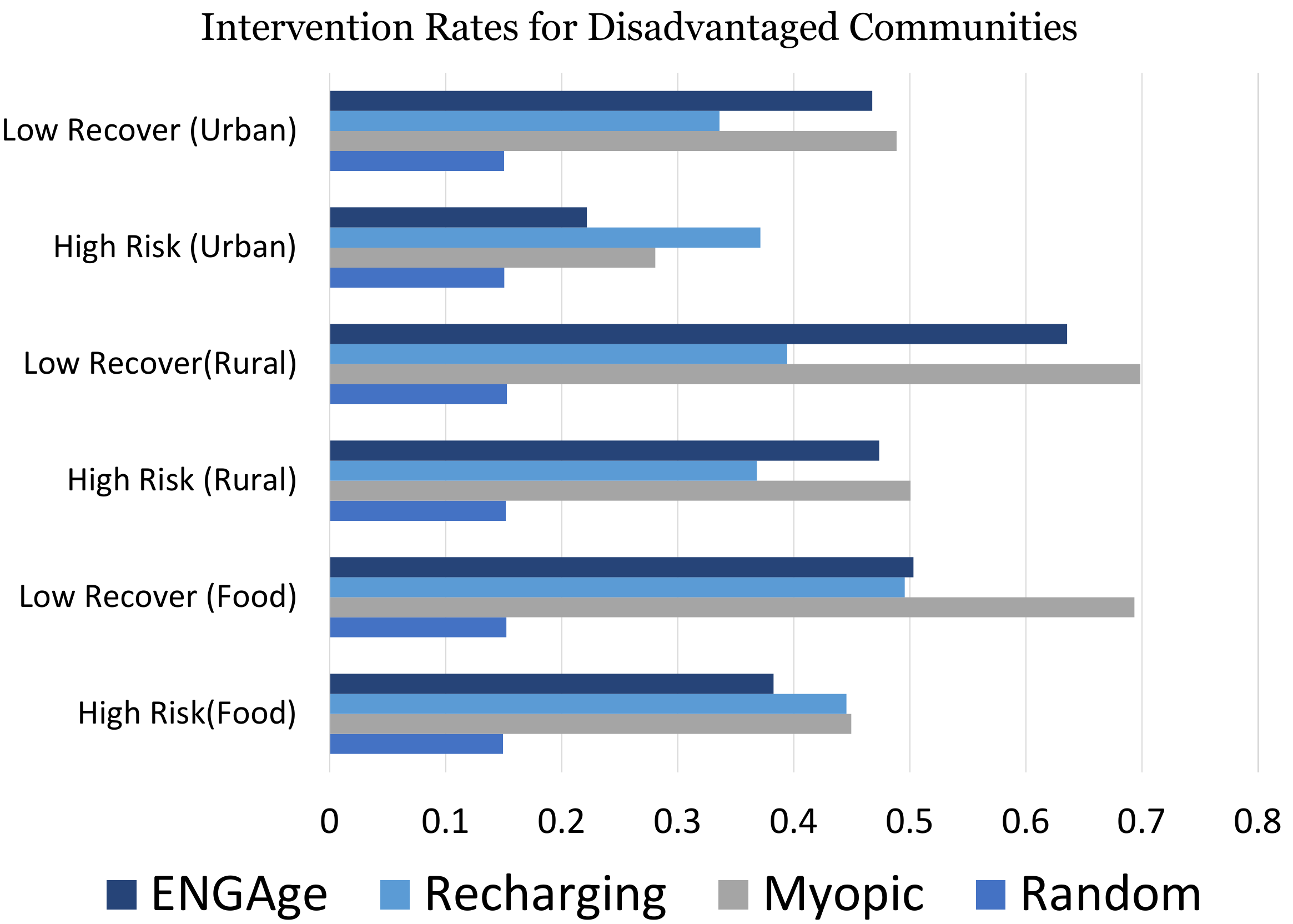}\label{fig:disadvantaged}%
}\hfill
\subfloat[Runtime]{%
  \includegraphics[width=0.3\textwidth,keepaspectratio]{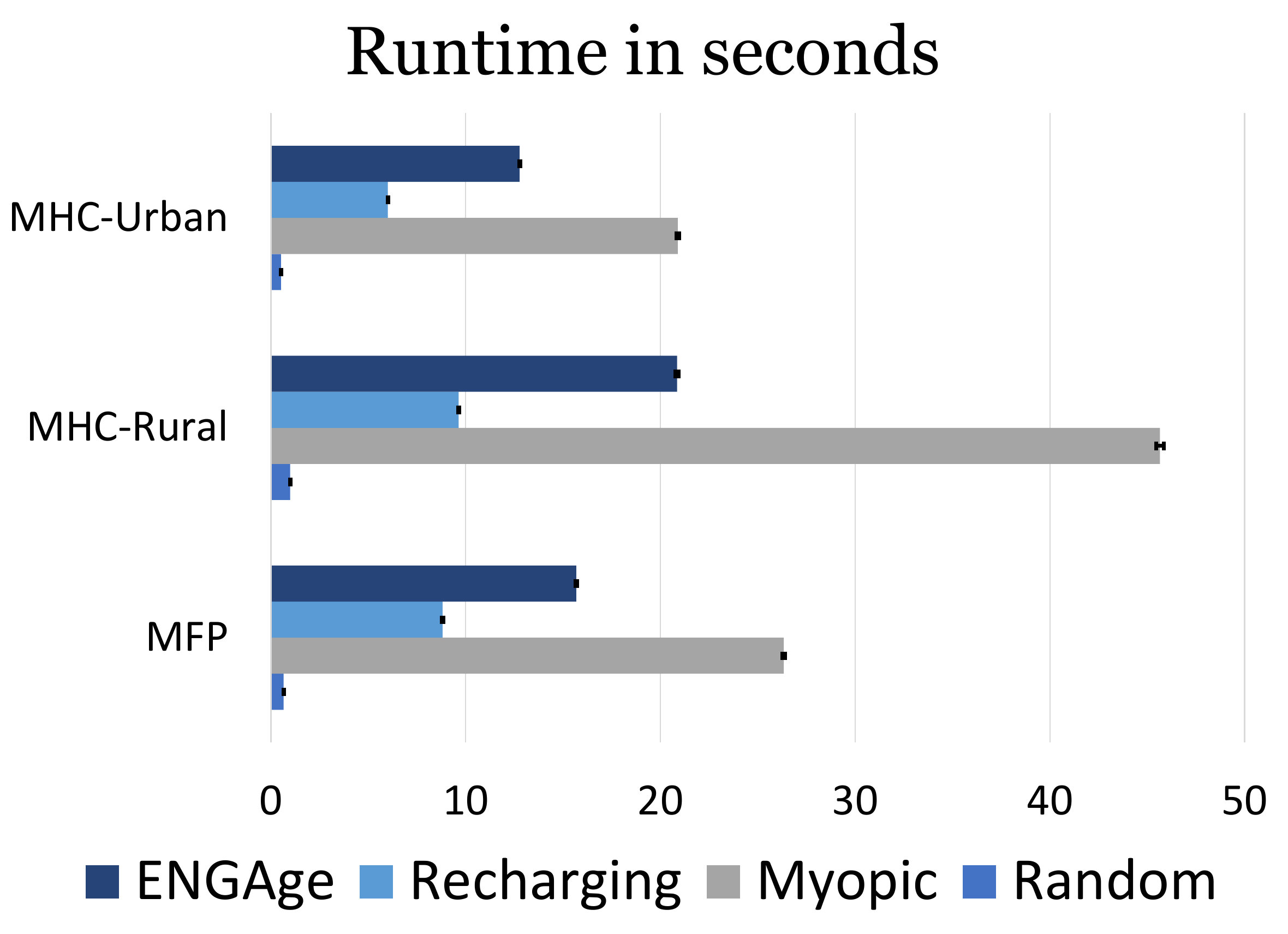}\label{fig:run_time}%
}\hfill
\centering 
\caption{(\ref{fig:average_degree}): average reward vs. graph average degree (\ref{fig:disadvantaged}): intervention rates for $15\%$ most disadvantaged communities. (\ref{fig:run_time}): average runtimes. }\label{fig:analysis}
\end{figure*}

\section{Experimental Evaluation}\label{sec:exp}
We perform experiments comparing our algorithm to baselines in a variety of real-world application scenarios. We begin by describing the application domains and their properties:

\textbf{Mobile Health Clinics in urban areas}: This domain setting is modeled on MHCs that are an important part of urban health care programs. Specifically, we consider a graph of the city of Boston (where such MHCs are used by non-profit organizations \cite{chen2022using}), collected from~\cite{JPXJTV_2017}. The graph consists of 431 locations that are used as bandit arms. The populations $n_v$ and transition probabilities $(p^p_{vGB},p^p_{vBG},p^a_{vBG},p^a_{vGB})$ are generated from uniformly random distributions subject to the assumptions introduced in the problem formulation section\footnote{While we have access to real-world street graph data, we do not have access to population and commuting data at a matching granularity.}.

\begin{table}
\caption{Properties of the network data sets. }
\scalebox{1}{\begin{tabular}{l|ccccc}
Network  & $|V|$ & average  & average degree \\
&& degree & centrality \\
\midrule\midrule
\textbf{Boston} & 431 & 2.92 & 0.005 \\
\textbf{Daniels County} & 631 & 2.53 & 0.008 \\
\textbf{Los Angeles} & 561 & 2.85 & 0.001\\
\end{tabular}}
\label{table:settingstable}
\end{table}

\textbf{Mobile Health Clinics in rural areas}: In contrast to urban areas, rural areas are characterized by a larger number of less connected smaller communities, and may experience lower overall levels of access to health services. We model this domain using a graph of Daniels County, MT, with 631 locations, taken from~\cite{JPXJTV_2017}.  Daniels County is considered one of the most rural counties in the US, as measured by the index of relative rurality \cite{Rurality}. We modify the previous setting to set a large portion of districts to have communities with relatively small population, to account for the characteristics described before. 


\textbf{Mobile Food Pantry}: 
Due to a limited choice of means of transportation, residents of many socially disadvantaged neighborhoods can only access food within shorter distances; as a result, healthy food options are often limited. Mobile food pantries (MFPs) have become an important source of healthy food for these communities~\cite{algert2006disparities}. In the MFP scenario, the Los Angeles city graph with 561 locations collected from~\cite{JPXJTV_2017} is used, as food insecurity is an important issue in Los Angeles. In this scenario, it is assumed that there is no prevention effect ($p^{a}_{GB}=p^{p}_{GB}$), as the provided food needs to be fresh and will only be distributed to individuals in bad states.
\begin{figure}[ht]
\centering
\includegraphics[width=0.4\textwidth,keepaspectratio]{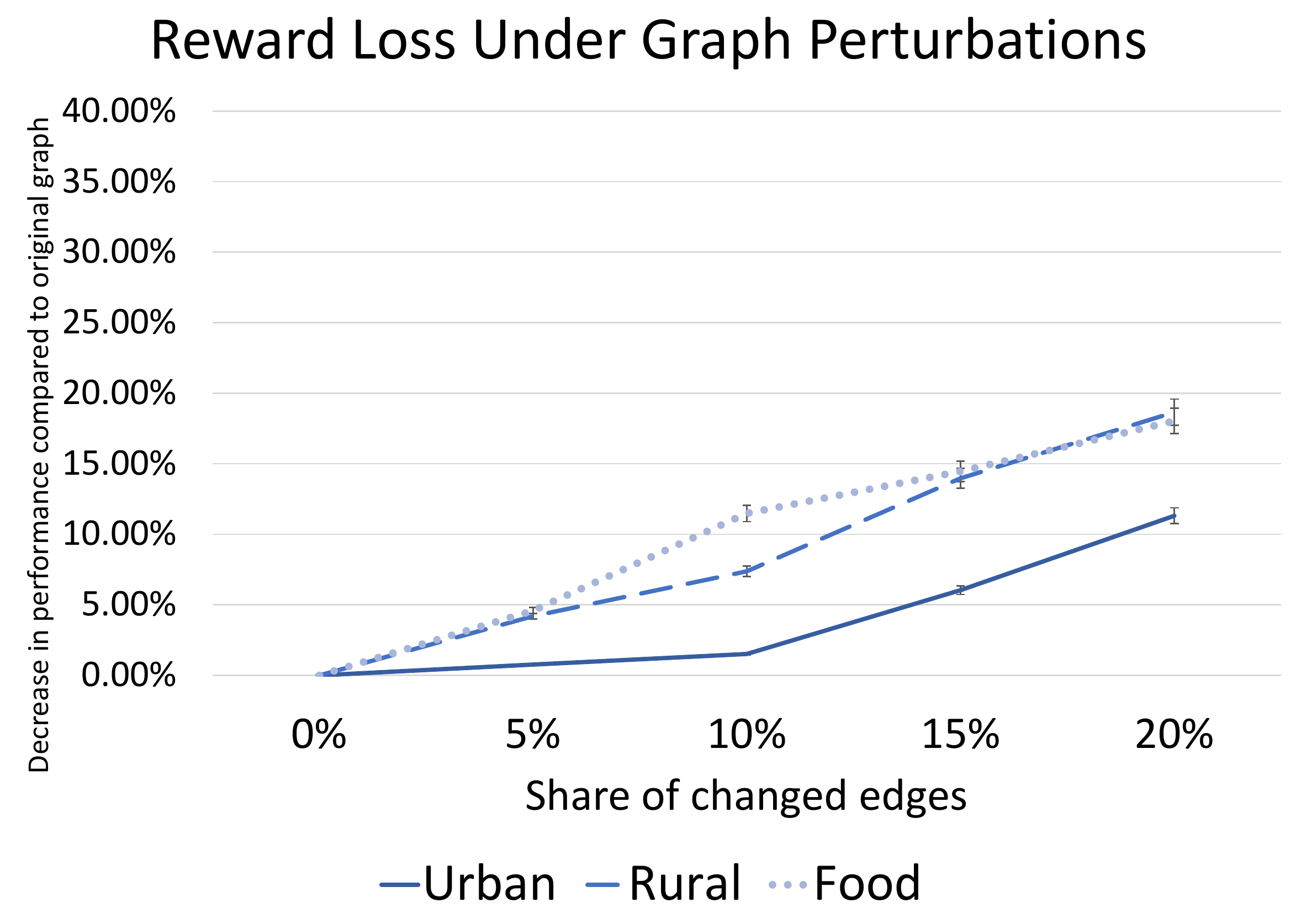}
\caption{Sensitivity analysis.} \label{fig:sensitivity}
\end{figure}



We compare our algorithm to three baseline algorithms. \textsc{Random} selects $k$ locations uniformly at random in each time step. \textsc{Myopic} selects the locations with maximum reward in the current time step. \textsc{Recharging} is the rounding scheme scheduling provided in~\cite{kleinberg2018recharging}. All experiments are conducted on a system with 6 cores, 2.60 GHz Intel CPU, and 16 GBs of RAM for $30$ simulations over $100$ time steps for each trial. All figures include approximate $95\%$ confidence intervals as error bars. Figures~\ref{fig:MHC}--\ref{fig:food} show the average reward collected with different budgets of $k\in\{10,20,30\}$ arms, for the three domains described above. Our algorithm consistently outperforms all baselines. 
\textsc{Recharging} mostly performs second-best, though in the urban MHC setting, it is slightly worse than \textsc{Random}.
Figure~\ref{fig:run_time} shows the average runtime per simulation in seconds. Interestingly, \textsc{Myopic} is the slowest algorithm, because it has to compute the reward for each node in each round, while \textsc{ENGAge} and \textsc{Recharging} use pre-computed period tables. 

We further analyze the sensitivity of these results to several modeling parameters. Figure~\ref{fig:average_degree} shows the performance of the algorithms for different densities for a synthetic domain based on a spatial preferential attachment model \cite{barthelemy2011spatial,ferretti2011preferential}. The results are non-monotonic for the \textsc{ENGAge} algorithm. A possible explanation could be that there might exist a level of optimum connectivity, below which adding more links will increase the intervention benefit by spreading interventions more widely, and above which adding more links will cause too much overlap between the populations that are intervened on in different time steps.
Figures~\ref{fig:cure_all} and \ref{fig:prevenrion_all} show that \textsc{ENGAge} consistently outperforms the baselines across multiple values for cure and prevention rates in all domains.

We also analyze the impact of our algorithm on the most disadvantaged communities, i.e., those experiencing the highest risk of transitioning to the bad state, or which have small probability of recovering from the bad state. Figure~\ref{fig:disadvantaged} shows the average intervention frequencies for the $15\%$ communities with the highest risk ($p^{p}_{GB}$) and lowest chance of recovery ($p^{p}_{BG}$). All algorithms except \textsc{Random} intervene on the most disadvantaged communities disproportionately more often, showing that they are not discriminating against them. This is thanks to the design of the reward criterion that measures intervention benefit for individuals receiving the intervention. 

Finally, we conduct a sensitivity analysis of the \textsc{ENGAge} algorithm against graph perturbations. Figure~\ref{fig:sensitivity} is constructed as follows: Starting with the real-world graphs from the three domains, we add perturbations by removing a given percentage of the edges, and adding back the same number of edges randomly. In the optimization, we then use the perturbed graph, while the original, unperturbed graph is used to compute the rewards. Overall, we observe that perturbing $x\%$ of edges generally reduces reward by less than $x\%$. For example, with a graph perturbation of $15\%$, the performance reductions in the urban, rural and food settings are $6\%$, $13\%$, and $14\%$, respectively.

\section{Conclusion}
We present a networked RMAB model motivated by mobile interventions; our model captures network effects stemming from traveling behavior. Our model was built based on the input of domain experts in mobile health interventions. To the best of our  knowledge, this is the first paper addressing the challenge of scheduling multiple interventions with network effects in the RMAB model. Network effects induce strong reward coupling between arms, substantially complicating the analysis of the RMAB. We propose the \textsc{ENGAge} (Efficient Network Geography Aware scheduling) algorithm that takes reward coupling and network effects into account. We provide sufficient conditions for optimality and show that our algorithm outperforms several baselines empirically in three real-world domains and
synthetic domains with varying properties. 




\begin{acks}
This work was supported by the Army Research Office (MURI W911NF1810208). J.A.K. was supported by an NSF Graduate Research Fellowship under Grant DGE1745303.
\end{acks}

\bibliographystyle{ACM-Reference-Format}
\balance
\bibliography{sample}


\end{document}